\documentclass[10pt,twocolumn,letterpaper]{article}

\usepackage{cvpr}              
\usepackage{graphicx}
\usepackage{amsmath}
\usepackage{amssymb}
\usepackage{xfrac}
\usepackage{color}
\usepackage{colortbl}
\usepackage{enumitem}
\usepackage[dvipsnames]{xcolor}
\usepackage{sidecap}
\usepackage{wrapfig}
\usepackage{lipsum} 
\usepackage{textcmds}
\usepackage{graphicx}
\usepackage{multirow}
\usepackage{booktabs}
\usepackage{colortbl}
\usepackage{xcolor}
\usepackage{hhline}
\usepackage{lipsum}
\usepackage{etoolbox}
\usepackage{capt-of,etoolbox}
\newcommand{\methodname}{SMILE}
\newcommand{\methodnamespace}{SMILE }

\usepackage{graphicx}
\usepackage{amssymb}
\usepackage{amsmath}
\usepackage{booktabs}
\usepackage{bbm}
\usepackage{times}
\usepackage{epsfig}
\usepackage{tabu}
\usepackage{multirow}
\usepackage{amsfonts}       
\usepackage{xpunctuate}
\usepackage{pifont}
\usepackage{color,colortbl}
\usepackage{tabulary,overpic}
\usepackage{booktabs}       
\usepackage[normalem]{ulem}

\usepackage{amsmath}
\usepackage{cite}

\usepackage{scalerel}
\usepackage[accsupp]{axessibility}

\usepackage{xcolor}
\usepackage{enumitem}
\usepackage{arydshln} 
\usepackage{xspace}
\usepackage{booktabs}
\usepackage{colortbl}
\usepackage{multirow}
\usepackage{makecell}
\usepackage{lipsum} 
\usepackage{gensymb} 

\usepackage{tabularx}

\definecolor{fyxcolor}{RGB}{0,128,255}
\definecolor{demphcolor}{RGB}{100,100,100}
\definecolor{citecolor}{RGB}{0,0,192} 
\definecolor{GrayBG}{gray}{0.95}

\newcommand{\app}{\raise.17ex\hbox{$\scriptstyle\sim$}}
\newlength\savewidth\newcommand\shline{\noalign{\global\savewidth\arrayrulewidth
  \global\arrayrulewidth 1pt}\hline\noalign{\global\arrayrulewidth\savewidth}}
\newcommand{\tablestyle}[2]{\setlength{\tabcolsep}{#1}\renewcommand{\arraystretch}{#2}\centering\footnotesize}

\newrobustcmd{\B}{\bfseries}

\usepackage[labelsep=period]{caption}
\renewcommand{\paragraph}[1]{\vspace{0.2em}\noindent \textbf{#1 \hspace{0.2em}}}
\definecolor{lightcyan}{rgb}{0.88, 1.0, 1.0}

\definecolor{MyDarkRed}{rgb}{0.66, 0.16, 0.16}
\definecolor{MyDarkBlue}{rgb}{0.16, 0.16, 0.66}

\definecolor{cvprblue}{rgb}{0.21,0.49,0.74}
\usepackage[pagebackref,breaklinks,colorlinks,allcolors=cvprblue]{hyperref}
\hypersetup{
    citecolor=green,
    linkcolor=red,
    urlcolor=magenta
}

\def\paperID{14630} 
\def\confName{CVPR}
\def\confYear{2025}

\title{
SMILE: Infusing Spatial and Motion Semantics in  Masked Video Learning  }

\author{Fida Mohammad Thoker, Letian Jiang, Chen Zhao$^\dag$, Bernard Ghanem \\
King Abdullah University of Science and Technology (KAUST) \\
{\tt\small \{fida.thoker,letian.jiang,chen.zhao,Bernard.Ghanem\}@kaust.edu.sa}
}

\begin{document}
\maketitle


\begin{abstract}
Masked video modeling,  such as VideoMAE, is an effective paradigm for video self-supervised learning (SSL). However, they are primarily based on reconstructing
pixel-level details on natural videos which have substantial temporal redundancy, limiting their capability for semantic representation and
sufficient encoding of motion dynamics.   To address these issues, this paper introduces a novel  SSL approach for video representation learning, dubbed as \methodname, 
by infusing both spatial and motion semantics. 
In \methodname, we leverage image-language pretrained models, such as CLIP, to guide the learning process with their high-level spatial semantics. We enhance the representation of motion by introducing synthetic motion patterns in the training data, allowing the model to capture more complex and dynamic content. Furthermore, using \methodname, we establish a new self-supervised video learning paradigm capable of learning strong video representations without requiring any natural video data.
We have carried out extensive experiments on 7 datasets with various downstream scenarios. \methodnamespace surpasses current state-of-the-art SSL methods, showcasing its effectiveness in learning more discriminative and generalizable video representations. Code is available: \color{magenta}{\url{https://github.com/fmthoker/SMILE}}
\end{abstract}    
\vspace{-5pt}
\section{Introduction}
\label{sec:intro}
\begin{figure}[t]
    \centering
    \captionsetup{font=small,skip=1mm}
    \includegraphics[width=0.95\linewidth]{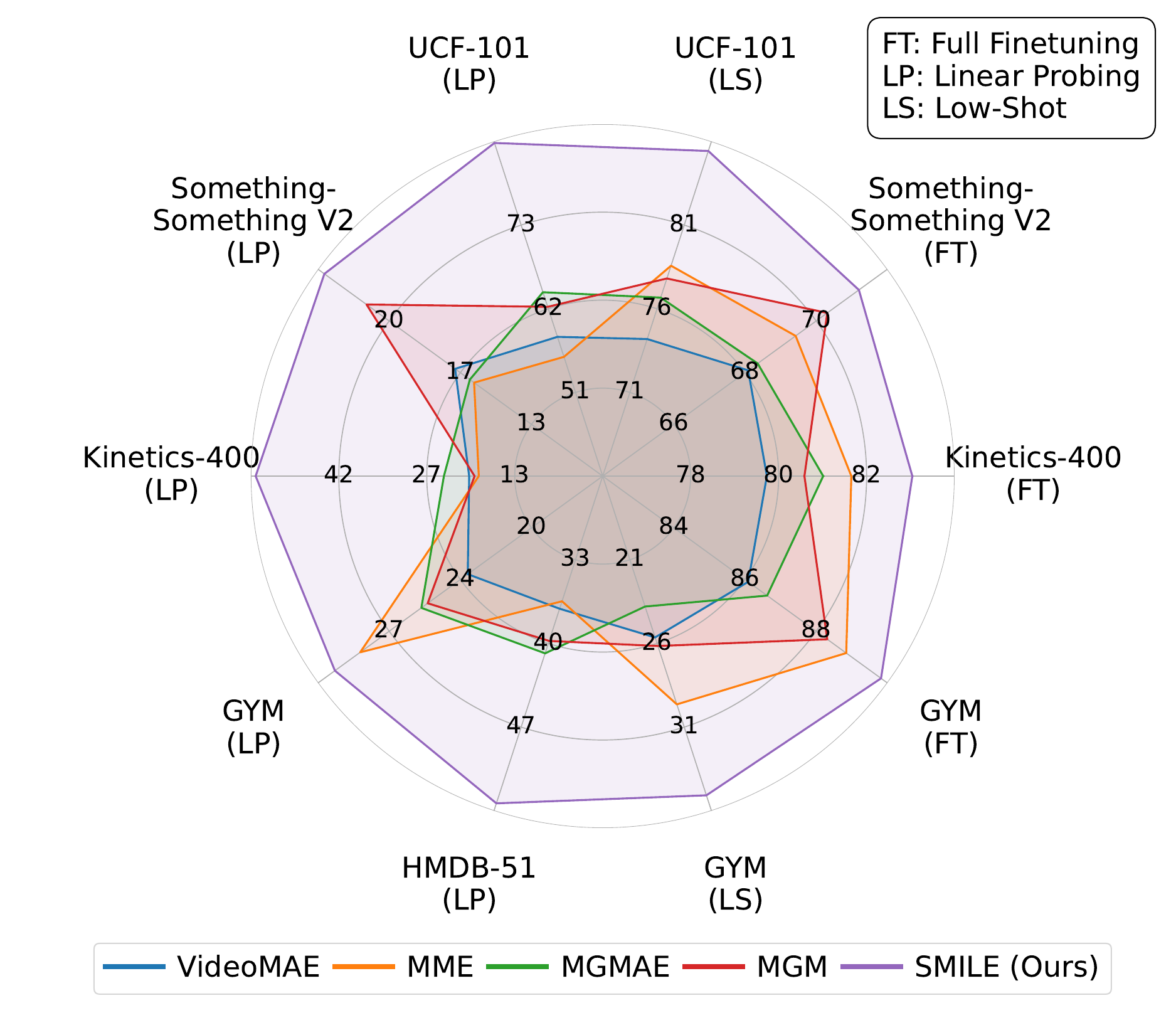}
    \vspace{-0.2cm}
    \caption{\textbf{Comparison with SOTA masked video modeling methods}. Our method significantly outperforms prior masked video modeling methods across diverse downstream settings.}
    \label{fig:teaser}
\end{figure}

Self-supervised learning (SSL)~\citep{wang2021unsupervised,qian2021spatiotemporal,piergiovanni2020evolving,tong2022videomae} plays a crucial role for many video understanding tasks, such as action recognition~\citep{Kinetics-400-arxiv,goyal2017something,UCF-101-arxiv,charades-sigurdsson:hal-01418216,thoker2022severe} and detection~\citep{yowo,liu2024end}, significantly reducing the reliance on extensive human annotation.  The goal of video SSL is to learn video representations that capture both spatial and temporal semantics effectively. Masked video modeling has emerged as a promising approach with state-of-the-art performance~\citep {tong2022videomae,fan2023mgm,mme_sun,huang2023mgmae}.
It involves masking a large portion of video frames and reconstructing the masked regions using the unmasked parts.

However, this approach has some limitations. \textit{First}, it relies heavily on pixel reconstruction, which can lead to overfitting low-level details. This limits the model’s ability to learn meaningful high-level semantics essential for discriminative representations, as shown by the suboptimal linear probing performance of current masked video modeling methods~\citep{tong2022videomae,huang2023mgmae,fan2023mgm,mme_sun} in Figure~\ref{fig:teaser}. The \textit{second} key issue in masked video modeling is the substantial temporal redundancy of video data with most videos having little foreground motion, human or object. 
Consequently, the reconstruction objective can be solved using the temporal correlations between co-located patches or by spatial correlations in tube masking~\citep{tong2022videomae}, resulting in inadequate motion awareness in the learned representations.
This inturn leads to poor generalization on downstream tasks, requiring a finer motion understanding as shown in~\citep{thoker2022severe}.

In this work, we address these limitations by proposing a novel framework for masked video modeling, dubbed as \methodnamespace (\textbf{S}patial and \textbf{M}otion semant\textbf{I}cs in masked video \textbf{LE}arning), to capture high-level spatial and motion semantics leading to generalizable video representations.
To improve the spatial semantics in masked video modeling, we propose to explicitly infuse high-level spatial semantics from pretrained image foundation models such as CLIP~\citep{radford2021learning}. Such models 
encapsulate high-level semantics learned through large-scale image-text alignment and have been adapted for video understanding tasks. For example, ActionCLIP~\citep{wang2021actionclip}, ViFiCLIP~\citep{rasheed2023fine}, and X-CLIP~\citep{ni2022expanding}) adapt CLIP  
for action recognition using videos with action labels, whereas ViCLIP~\citep{wang2023internvid} and VideoPrism~\citep{zhao2024videoprism}) adapt it for video-text alignment with video-text pairs. 
We take inspiration from them but aim to learn video representations with unlabeled videos. 
In particular, we propose replacing the reconstruction of pixels in masked video modeling with that of CLIP embedding, thus providing high-level semantics as the supervision signal. 

To improve motion awareness, we propose to reduce the temporal redundancy in input training videos with synthetic motion patterns.
Specifically, we randomly paste objects onto the video frames and move them smoothly along the temporal dimension to simulate variable temporal transitions of motion-intensive videos. This reduces the temporal redundancy in the original videos and results in videos with an additional prominent foreground motion.  
Furthermore, adding objects with known locations allows us to add a motion focus to the mask-and-predict task by explicitly masking the regions of object motion. This in turn encourages the network to avoid an over-reliance on static spatial information and encode the motion dynamics of the added objects, when solving the reconstruction task. 
We summarize our contributions as follows.
\begin{itemize}
\item We introduce \methodname, a novel SSL approach for masked video modeling.  It encodes spatial and motion semantics using guidance from image-text pretrained models and synthetic-motion augmented videos, leading to more discriminative and generalizable video representation.
\item We present a new strategy to augment video data with diverse synthetic motions, paired with a masking strategy that focuses on reconstructing these added motions to improve masked video modeling pretraining.
\item Through extensive evaluation on 7 datasets, including Kinetics-400~\citep{Kinetics-400-arxiv}, UCF-101~\citep{UCF-101-arxiv}, HMDB-51~\citep{kuehne2011hmdb}, FineGYM~\citep{shao2020finegym}, EPIC-Kitchens-100~\citep{EPIC-100-arxiv}, SomethingSomething V2~\citep{goyal2017something}, and Charades~\citep{charades-sigurdsson:hal-01418216}, we demonstrate that \methodname\ significantly outperforms existing state-of-the-art video SSL methods and exhibits better generalization.
\end{itemize}
\vspace{-0.2cm}
\section{Related Works}
\label{sec:formatting}
\vspace{-0.1cm}
\noindent\textbf{Video Self-Supervised Learning.} Video SSL aims to learn the spatio-temporal dynamics of video data without the need for any supervision. Three main learning paradigms have emerged: transformation prediction,  contrastive learning, and masked video modeling. 
In transformation prediction, models learn video patterns by solving tasks such as space-time puzzles ~\citep{kim2019self,videojigsaw,3drotnet}, predicting order in video clips~\citep{misra2016shuffle,xu2019self,sorting,odd,vcp,simon}, or estimating speed~\citep{benaim2020speednet, cho2020self,prp}. 
Contrastive learning is based on instance discrimination~\citep{infonce} and learning invariances to spatio-temporal augmentations~\citep{videomoco-pan2021videomoco,qian2021spatiotemporal,svt,gdt-patrick2020multimodal,chen2021rspnet,large-scale-feichtenhofer2021large,alwassel_2020_xdc,thoker2023tubelet,avid-cma-morgado2021audio,thoker2021skeleton}.  Masked video modeling methods~\citep{tong2022videomae,fan2023mgm,mme_sun,huang2023mgmae,yang2022motionmae}, like VideoMAE~\citep{tong2022videomae}, use a mask-and-predict task to reconstruct videos from partial data with an encoder-decoder setup similar to MAE~\citep{he2022masked}. By masking a high proportion of pixels (\textit{e.g.}, 90\%) and reconstructing them such methods learn useful video representations.  Our approach aims to improve the spatial and motion semantics of the video representations learned by masked video modeling. 

\noindent\textbf{CLIP for Video Understanding.}
CLIP~\citep{radford2021learning}  captures detailed spatial semantics from extensive image-text training, making it a strong foundation for directly modeling video data with added temporal dynamics, \textit{e.g.},~\citep{wang2023internvid,wang2021actionclip,ni2022expanding,park2023dualpathadaptationimagevideo,yang2023aimadaptingimagemodels,tu2023implicittemporalmodelinglearnable}.  ViFiCLIP~\citep{rasheed2023fine} finetunes CLIP on video by adding a temporal pooling layer, ActionCLIP~\citep{wang2021actionclip} includes a temporal positional embedding and aggregation layer, and X-CLIP~\citep{ni2022expanding} integrates cross-frame attention to share information between frames.  
Most relevant to us are works that also use CLIP for video representation learning~\citep{wang2023internvid,li2023unmasked,fan2024text}. ViCLIP~\citep{wang2023internvid} initializes a model with CLIP weights and employs further contrastive pretraining with 
video-text pairs.
UMT~\citep{li2023unmasked} uses CLIP as a teacher and aligns its predictions with the CLIP features for unmasked tokens, followed by a 2nd stage of learning from video-text contrastive pairs. TGM~\citep{fan2024text} relies on CLIP text-encoder features to guide masked video modeling and video-text contrastive learning in a unified framework.
In contrast, our \methodnamespace only learns from video data in a self-supervised manner, reconstructing CLIP features of masked tokens and added objects.

\noindent\textbf{Motion-aware Video Self-Supervised  Learning.} 
Encoding motion dynamics in video representations is essential for generalizing across various video tasks~\citep{thoker2022severe}. As a result, many video SSL methods
focus on enhancing the motion sensitivity of learned representations. For example,~\citep{motion_fit, coclr, mscl, xiao2022maclr,huang2023mgmae} capture motion using optical flow, ~\citep{background_removing, fame} do so by removing background or static elements and ~\citep{motion_static, wang2021unsupervised,thoker2023tubelet,doughty2024locomotion} incorporate synthetic motions.
\citep{motion_static, wang2021unsupervised} learn by predicting the speed and trajectory of the pseudo motions, and ~\citep{thoker2023tubelet} contrasts video with different types of motions. 
These approaches are specific to transformation prediction and contrastive learning paradigms. We take inspiration from these works and inject synthetic motions into the masked video learning.

Most relevant to our work are methods like~\citep{huang2023mgmae,fan2023mgm,yang2022motionmae,mme_sun}, which enhance motion understanding in VideoMAE~\citep{tong2022videomae} by modifying certain design choices. MGMAE~\citep{huang2023mgmae} and MGM~\citep{fan2023mgm} improve masking strategies to mask the regions of motion using optical flow and motion vectors. MotionMAE~\citep{yang2022motionmae} adds frame difference as an additional target for reconstruction, and MME~\citep{mme_sun} uses optical flow to extract dense object trajectories and then predicts these trajectories and their HOG features. However, these methods are limited by the temporal redundancies in the original data, restricting the amount of motion or motion-focused regions available for learning in the reconstruction task. In contrast, we add synthetic motions to the original video data, reducing temporal redundancy and increasing motion-rich regions to learn from. As shown in our experiments, reconstructing these modified videos encourages the model to capture motion dynamics, enhancing the motion awareness of the learned video representations.

\begin{figure*}[t]
    \vspace{-0.5cm}
    \centering
    \captionsetup{font=small,skip=1mm}
    \includegraphics[width=0.95\linewidth]{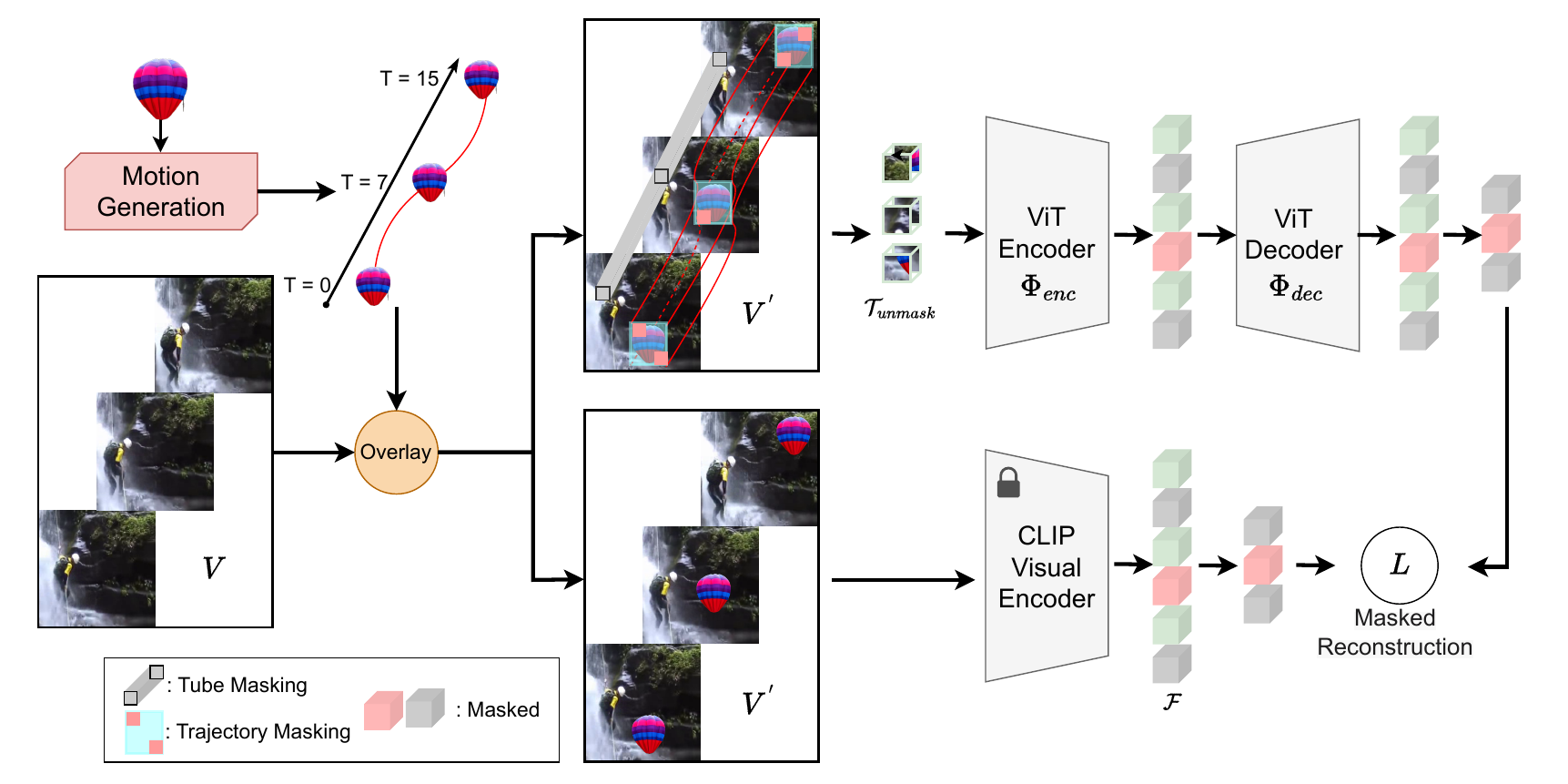}
    \vspace{-0.2cm}
    \caption{\textbf{Overall architecture of our \methodname.} An input video clip $V$ is overlaid with a segmented object along a randomly generated trajectory to generate $V^{'}$ infusing synthetic object motion in $V$. $V^{'}$ is passed frame-by-frame through the CLIP encoder to extract feature tokens $\mathcal{F}$. $V^{'}$ is patchified into a set of space-time tokens $\mathcal{T}$. We apply two types of masking upon $\mathcal{T}$, tube masking on the space-time tokens of the original video and trajectory-based masking on the tokens of added objects. The unmasked tokens $\mathcal{T}_{unmask}$ are fed into the encoder-decoder network $\Phi_{enc}$$\Phi_{dec}$ which is trained to reconstruct the masked feature tokens in $\mathcal{F}$.  
    }
    \label{fig:main}
\end{figure*}

\vspace{-0.2cm}
\section{Methodology}

\subsection{Revisiting Masked Video Modeling} \label{subsec:revisiting_videomae}
\vspace{-0.1cm}

The goal of masked video modeling~\citep{tong2022videomae,huang2023mgmae,fan2023mgm} is to learn video representations by employing a mask-and-predict task with an asymmetric encoder-decoder architecture. An input video $V \in \mathbb{R}^{T \times H \times W \times 3}$  is first divided into non-overlapped 3D space-time tokens (\textit{e.g.}, $2\times16\times16$)  as \(\mathcal{T}\) $=$ \(\{v_1, \dots, v_N\}\). From  \(\mathcal{T}\), a subset of tokens \(\mathcal{T}^{mask}\) are randomly masked with a high masking ratio (\textit{e.g.}, 90\%) and the unmasked tokens \( \mathcal{T}^{unmask} = \mathcal{T} \setminus \mathcal{T}^{mask} \) are fed into a transformer encoder $\Phi_{enc}$. Then, the output of the encoder and the learnable tokens containing the positional information of the masked tokens are passed to a shallow decoder $\Phi_{dec}$ to reconstruct the input  pixel with the following loss:
\begin{eqnarray}
    \mathcal{L} = \frac{1}{\lvert \mathcal{T}^{mask}\ \rvert} \sum_{i \in \mathcal{T}^{mask}} \lVert {v}_{i} - Y_{i} \rVert_2^2,
\label{eq:reconstruction_loss}
\end{eqnarray}
where $v_i$ refers to the \textit{pixel values} of the  $i$-th space-time token in $\mathcal{T}^{mask}$ and $Y_{i} = \Phi_{dec}(\Phi_{enc}(\mathcal{T}^{unmask}))$  denotes the corresponding reconstructed pixels for $i$-th token given the set of all unmasked tokens $\mathcal{T}^{unmask}$. 

Such pixel reconstruction drives the model to focus on low-level features, and with high temporal redundancy in video data, it primarily reconstructs static appearance details of video data. However, effective video understanding also requires capturing temporal progression and high-level semantics, such as scene context and object interactions.

\subsection{General Scheme of \methodname} \label{subsec:overview}
\vspace{-0.1cm}
Our goal is to infuse spatial and motion semantics in masked video modeling. To achieve this, we integrate two key elements into the masked video modeling paradigm. 
To improve motion semantics, we overlay local motion patterns in input videos in the form of moving objects.  To improve spatial semantics, we replace the low-level pixel targets with high-level features extracted from a pretrained CLIP encoder.  By combining both, we learn video representations that can encode both spatial and motion dynamics necessary for modeling downstream video tasks.  The overview of our method is shown in Figure~\ref{fig:main}.

\noindent\textbf{Motion Infusion.} Given a input video  $V\in\mathbb{R}^{T\times H\times W \times 3}$. We first sample a random object $o$ from a set of predefined segmented objects $O$. 
Next, we generate a motion trajectory (Section \ref{subsec:motion_generation}), along which we overlay this object onto the video $V$ to obtain the motion-augmented video $V'$. By adding objects with synthetic motions, we reduce temporal redundancy in the original videos and encourage the model to learn from the motion dynamics of the overlaid objects.

\noindent\textbf{Masking.}  We employ two separate masking strategies on $V'$ as follows. For the original regions, we employ random tube masking as in VideoMAE~\citep{tong2022videomae}. 
For added object regions,  we propose \textit{trajectory-based masking}. Since we know the location of the object in each frame of $V'$, we explicitly drop space-time tokens at random along its trajectory. Employing a high masking ratio along the object trajectory ensures that solving the reconstruction task requires encoding the object\textquotesingle s motion dynamics, thereby enhancing the motion awareness of learned representations.

\noindent\textbf{Architecture.}
We employ a teacher-student reconstruction architecture with a pretrained image-encoder as the teacher and an asymmetric encoder-decoder network $\Phi$ as the student.  The augmented video is patchified into 3D space-time tokens \(\mathcal{T}\) $=$ \(\{v^{'}_{1}, \dots, v^{'}_{N}\}\) as in general video masked autoencoders (Section~\ref{subsec:revisiting_videomae}). The tokens are then masked using tube and trajectory-based masking to generate sets of masked $\mathcal{T}^{mask}$ and unmasked $\mathcal{T}^{unmask}$ tokens. The unmasked tokens $\mathcal{T}^{unmask}$ and the learnable tokens containing the positional information of the masked patches are used as the input to $\Phi$ as in Section~\ref{subsec:revisiting_videomae}.

\noindent\textbf{Reconstruction.} 
Instead of directly reconstructing pixels, we project the input video $V'$  onto the teacher's feature space to generate high-level targets. In particular, we choose a CLIP model as the projection network due to its high-level semantics learned from language guidance. Specifically, the input $V'$  is passed frame-by-frame through the CLIP encoder to extract features.  
The extracted features are stacked 
into a sequence of feature tokens as \(\mathcal{F}\) $=$ \(\{f^{'}_{1}, \dots, f^{'}_{N}\}\). 
To match the number of tokens with  $\mathcal{T}$, we only use the CLIP features for the first time slice of the 3D space-time tokens, thus generating a total of $N$  feature tokens, each corresponding to a space-time token in the input video. 
The encoder-decoder is then trained with a masked feature reconstruction loss as:
\begin{eqnarray}
    \mathcal{L} = \frac{1}{\lvert \mathcal{T}^{mask}\ \rvert} \sum_{i \in \mathcal{T}^{mask}} \lVert f^{'}_{i} - Y_{i} \rVert_2^2,
\label{eq:reconstruction_loss_ours}
\end{eqnarray}
where $f^{'}_{i}$ is the \textit{CLIP feature projection} of $i$-th space-time token $v^{'}_{i}$ in $\mathcal{T}^{mask}$ and $Y_{i} = \Phi_{dec}(\Phi_{enc}(\mathcal{T}^{unmask}))$ denotes the corresponding reconstructed feature for $i$-th token given the set of all unmasked tokens $\mathcal{T}^{unmask}$.  Such formulation first increases the temporal dynamics in otherwise static data, avoiding learning shortcuts for reconstruction.  Second, projecting targets onto the CLIP space encourages the network to capture more high-level video semantics. 

Overall, the key intuition is that CLIP features capture high-level information, such as scene context and object interactions, which guides the network to attend to both the overlaid object motions and the original video semantics.

\vspace{-0.1cm}
\subsection{Motion Generation} \label{subsec:motion_generation}
\vspace{-0.1cm}
To infuse motion,  we overlay a segmented object along a randomly generated trajectory onto the original video frames. To introduce more complex dynamics, the object is also transformed \textit{e.g.} rotation, along the trajectory. 

\noindent\textbf{Appearance and Trajectory.} To infuse a new synthetic motion onto a video sequence $V{=}[v_1, v_2, ..., v_T]$ comprising $T$ frames, we begin by selecting an object $o$ from $O$ at random, where $O$ denotes the set of off-the-shelf segmented objects generated using Stable Diffusion~\citep{rombach2022high} and X-Paste\citep{zhao2023x}. We then sample a scale $P{\times}Q$ from a predefined set of sizes for $o$ to model different object sizes, where $P \ll H$ and $Q \ll W$. 
Next, we sample a sequence of locations for the object   $\mathrm{Traj}_{o} = [(x^1,y^1),..,(x^T,y^T)]$ representing the center coordinate $(x^i,y^i)$  of the object $o$ in the corresponding frame. Specifically, we generate a random non-linear path for the trajectory by sampling $M$ 2D points ($M \gg T$) uniformly from $x \in [0, H]$ and $y \in [0, W]$. To ensure smoothness in the motion,  a  $1D$ Gaussian filter is applied along $x$ and $y$ dimensions as $[(h(x^1), h(y^1)), \dots, (h(x^M), h(y^M))]$, where $h(z) = \frac{1}{\sqrt{2 \pi} \kappa} e^{-z^2 / 2 \kappa^2}$ and $\kappa$ is the  smoothing factor. The resulting trajectory is then downsampled to $T$ target locations for the object $o$. 
This formulation introduces diverse motions with varying spatio-temporal dynamics in the form of objects with different shapes, sizes, and temporal progression.

\noindent\textbf{Transformation.} To increase the complexity of the object's motion, we also apply scale and rotation transformations along the temporal dimension.  Specifically, we select 3 frame indices, the first, the last, and a random middle one, $v_1$,  $v_m$, and $v_T$, respectively. For these three frame indices, a rotation angle $\alpha_i$, and a scaling factor $S_i$ are selected randomly.    
The angles and scales for the rest of the frames are linearly interpolated between these keyframes.  Finally, the object $o$ is  rotated and scaled with the corresponding transformations $\big(\begin{smallmatrix} \cos \alpha_i & -\sin \alpha_i \\ \sin \alpha_i & \cos \alpha_i \end{smallmatrix}\big)$ and $\big(\begin{smallmatrix} S_{i} & 0 \\ 0 & S_{i} \end{smallmatrix}\big)$,  before overlaying it onto the video frame $v_i$.

\begin{figure}[t]
    \vspace{-0.4cm}
    \centering
    \captionsetup{font=small,skip=1mm}
\includegraphics[width=0.95\linewidth]{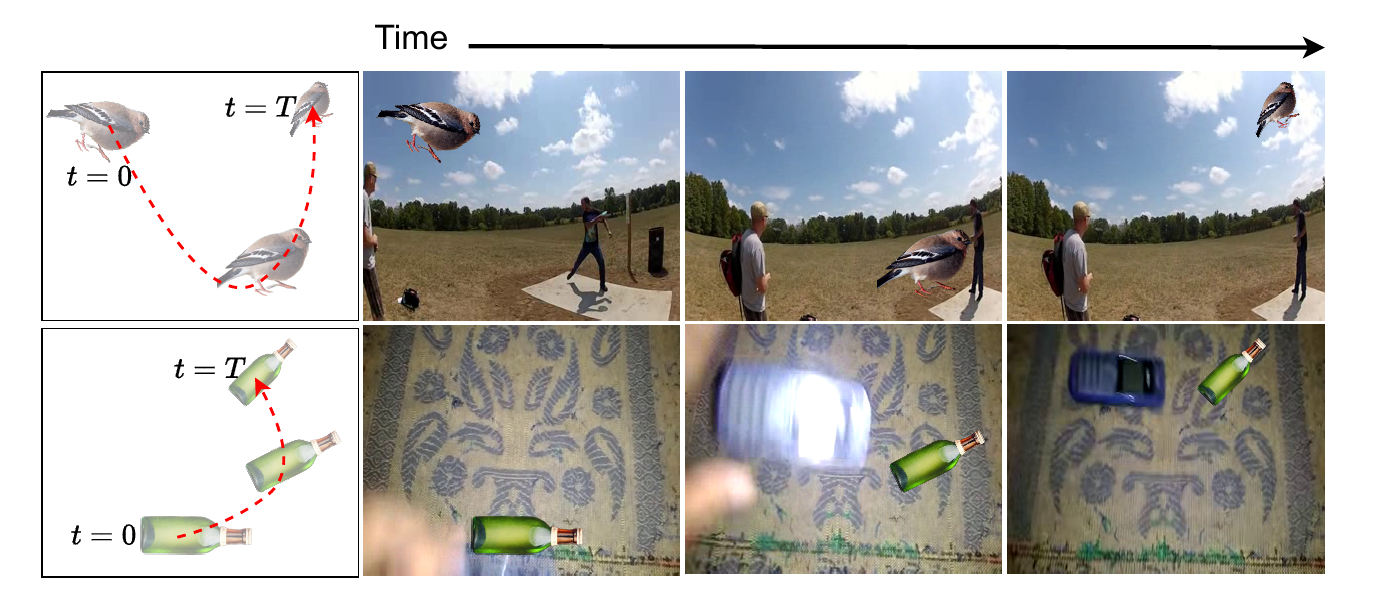}
    \caption{\textbf{Motion overlaid videos} showing the position change and transformation of the added objects along the time dimension. 
    } 
    \label{fig:motions}
\end{figure}

Figure~\ref{fig:motions} shows two sample videos with our added motions. While these motions may not appear entirely realistic, they represent complex motion dynamics of moving structured pixels, and introducing them into a mask-and-predict framework guides the model to learn from such dynamics. This results in video representations with improved downstream performance on motion-aware tasks, as we also validate empirically. Interestingly, we also show in the experiments (Section~\ref{sec:discussion}) that overlaying such object motions on black or noise frames can learn strong video representations without requiring any original videos.

\vspace{-.2cm}
\subsection{Training Procedure} 
\vspace{-.1cm}
By adding synthetic motions, we have introduced a domain shift in the input for reconstruction, which may affect the performance of downstream tasks.
To explore effective learning from both original videos ${V}$ and motion-overlaid videos  ${V'}$, we explore two training strategies, \textit{i.e.} \textbf{mixed training} and \textbf{progressive training.} In the \textit{mixed training} strategy, we use both ${V}$ and ${V'}$ as the input to the network and directly optimize for the reconstruction of both inputs. The idea is to simultaneously encode the spatio-temporal dynamics of the original video data and the motion dynamics of the augmented data, with a limited domain shift. In the \textit{progressive training}  strategy, we employ a two-stage learning approach that first learns from augmented data ${V'}$ alone, then continues to train the model with the original data ${V}$ alone.  This way we can first learn to encode the spatio-temporal dynamics of the augmented data and then align the network to further encode the dynamics of the original data and its domain. 

\vspace{-0.1cm}
\section{Experiments}
\label{sec:experiments}
\vspace{-0.2cm}
We evaluate \methodnamespace against prior video SSL methods on action recognition benchmarks: Kinetics-400 (K400)~\citep{Kinetics-400-arxiv}, SomethingSomething V2~\citep{goyal2017something} (SSv2), {UCF-101}~\citep{UCF-101-arxiv} (UCF), {HMDB-51}~\citep{kuehne2011hmdb} (HMDB), FineGYM~\citep{shao2020finegym} (GYM), and EPIC-Kitchens-100~\citep{EPIC-100-arxiv} (EPIC), achieving the best performance in linear probing and full finetuning (Section~\ref{sec:sota_comparison}). \methodnamespace also shows better generalization on the SEVERE benchmark\citep{thoker2022severe} (Section~\ref{sec:severe}). Ablations and further discussion are provided in Section~\ref{sec:ablation} and Section~\ref{sec:discussion}, respectively.

\vspace{-0.1cm}
\subsection{Implementation Details}
\vspace{-0.1cm}
\noindent\textbf{Model.} Following VideoMAE~\citep{tong2022videomae}, we use ViT-S or ViT-B with joint space-time attention for our student network. We choose CLIP ViT-B~\citep{radford2021learning} as our teacher network. \\
\noindent\textbf{Motions.} 
We follow PIN~\citep{dorkenwald2024pin} to create the set of segmented objects $O$ with Stable Diffusion~\citep{rombach2022high} and X-Paste~\citep{zhao2023x}.
We sample the object size $P{\times}Q$  uniformly from [$32{\times32}$, $128{\times}128$] and sample the rotation 
 angles from $[-90, 90]$ and scaling factors  from $[0.5,1.5]$.
 
\noindent\textbf{Pretraining.} We follow previous masked video modeling works~\citep{tong2022videomae,mme_sun,fan2023mgm,huang2023mgmae} and use Kinetics-400~\citep{Kinetics-400-arxiv} (K400) or SomethingSomething V2~\citep{goyal2017something} (SSv2)  for pretraining unless specified otherwise. We use the same hyperparameters as in VideoMAE~\citep{tong2022videomae}. Unless stated otherwise, we train with our proposed  \textit{progressive training} for a total of 600 epochs, 300 epochs in each stage. More training details are in  \textbf{supplementary} material.

\vspace{-0.1cm}
\subsection{Comparison with State-of-the-Art} \label{sec:sota_comparison}
\vspace{-0.2cm}
\paragraph{Linear Probing.} 
We freeze the pretrained backbone and only train a linear classifier on the target dataset.
When evaluating the prior works for comparison, we utilize their publicly available checkpoints for the ViT-B backbone pretrained on the K400 dataset for 800 epochs except for MVD (2000 epochs) and MME (1600 epochs). We evaluate the downstream task using a common setup for a fair comparison. 
Details about datasets and experimental settings are in the \textbf{supplementary} material.

\begin{table}[t]
    \centering
    \caption{\textbf{Linear probing comparison of masked video modeling methods.}  Our \methodnamespace learns highly discriminative video representations useful for both spatial and motion-focused domains. All results are obtained with the ViT-B backbone pretrained on K400 and Top-1 accuracies (\%) are reported.
    }
    \setlength{\tabcolsep}{0.30em}
    \small
    \begin{tabular}{lccccccc} \toprule \textbf{Method} & K400 &  UCF & HMDB & SSv2 &GYM & EPIC \\
         \midrule
         VideoMAE \citep{tong2022videomae} &  20.7 & 58.6 & 37.7 & 17.5 & 23.9 & {33.2} \\
         MVD \citep{mvd_wang}      &  18.7 & 49.1 & 28.6 & 12.2 & 22.7 & 29.7\\
         MME \citep{mme_sun}       &  19.1 & 56.0 & 37.1 & 16.6 & {29.0} & 32.2\\
         MGMAE \citep{huang2023mgmae} &  24.9 & 64.4 & 41.3 & 16.8 & 26.1 & {33.2} \\
         EVEREST \citep{hwangeverest}  &  14.1 & 51.8 & 30.3 & 14.5 & 23.3 & 30.5 \\
         MGM \citep{fan2023mgm} &  19.8 & 62.5 & 40.3 & \underline{21.7} & 25.8 & 32.4 \\
         SIGMA~\citep{salehi2025sigma}  &{47.5} & 80.7 & \underline{52.3} & 20.8 & \underline{30.1} & \underline{34.2} \\
        \rowcolor{lightcyan}
         \methodnamespace w/o motion           & \underline{54.6} & \underline{82.6} & {51.7} & 21.6 & 27.9 & 32.3 \\
        \rowcolor{lightcyan}
         \methodnamespace  (ours)            & \textbf{56.2} & \textbf{83.8} & \textbf{53.4} & \textbf{23.7} & \textbf{30.2} & \textbf{34.4} \\
        \bottomrule
    \end{tabular}
    \label{tab:linear-probing}
\end{table}
Table~\ref{tab:linear-probing} shows that our method, even without synthetic motion, significantly outperforms all state-of-the-art methods on K400 and UCF, and is among the top performers on HMDB, SSv2, GYM, and EPIC. This highlights that CLIP feature reconstruction offers a stronger supervisory signal than pixel reconstruction, especially in tasks requiring spatial semantics like K400 and UCF. Adding synthetic motions further boosts performance by an average of $1.8\%$ across datasets, enhancing motion cues through reconstruction. On motion-sensitive datasets like SSv2, GYM, and EPIC, our approach consistently surpasses MGM~\citep{fan2023mgm}, MGMAE~\citep{huang2023mgmae}, and MME~\citep{mme_sun}, confirming that synthetic motions help overcome the lack of motion diversity in current datasets that may limit these methods.

\paragraph{Full Finetuning.} We finetune both the pretrained backbone and the classification head end-to-end following \citep{tong2022videomae,mme_sun,huang2023mgmae} on the downstream datasets K400 and SSv2. Following prior masked video modeling works~\citep{tong2022videomae,fan2023mgm,huang2023mgmae}, we show results with self-supervised pretraining on K400 and SSv2 datasets for the VIT-B backbone. We follow VideoMAE~\citep{tong2022videomae} for finetuning and evaluation protocols with more details in the \textbf{supplementary}. 

Tables~\ref{tab:ssv2} and~\ref{tab:k400} show the results for SSv2 and K400, respectively. On SSv2, we achieve the best performance with both K400 and SSv2 pretraining, highlighting our method’s effectiveness for both in-domain and cross-domain transfer. For K400, we exceed the prior top SSL approach by 1.6\%. Additionally, our gains over motion-focused methods, such as MGMAE (1.5\% on SSv2, 1.9\% on K400), MME (2.5\% on SSv2, 1.6\% on K400), and MGM (1.9\% on SSv2, 2.3\% on K400), confirm our improved motion sensitivity, even in full-finetuning settings. Notably, our model surpasses VideoMAE, MotionMAE, MME, and MGMAE trained for more than twice the number of epochs (see \textbf{supplementary} for the extensive comparison). We attribute this efficiency to CLIP feature reconstruction, which directly captures high-level semantics compared to pixel-level learning. 
\begin{table}[t!]
 \vspace{-0.2cm}
\centering
\captionsetup[sub]{font=normalsize}
\caption{\textbf{Full finetuning comparison of various self-supervised methods on  Something-Something V2}. All results are obtained with a ViT-B backbone. $*$ denotes results obtained by our evaluation. Our \methodnamespace achieves state-of-the-art performance. Top-1 accuracies (\%) are reported.}
\tablestyle{2.5pt}{1.02}
\small
\begin{tabular}{l|c|c|c|c}
    \toprule
    \textbf{Method} & \textbf{Backbone} & \textbf{Epochs} & \textbf{Pretrain} & \textbf{Top-1} \\ 
    \shline 
    VideoMAE~\citep{tong2022videomae} & ViT-B & 800& K400 &68.5  \\
    OmniMAE ~\citep{girdhar2023omnimae} & ViT-B  & 800 & K400 &69.0 \\
    MGMAE$^*$~\citep{huang2023mgmae} & ViT-B & 800& {K400} &68.9   \\
    MME~\citep{mme_sun} & ViT-B & 800& {K400} &70.5  \\
    SIGMA~\citep{salehi2025sigma} & ViT-B & 800 & {K400} &71.1  \\
    MGM$^*$~\citep{fan2023mgm} & ViT-B & 800& {K400} &71.1  \\
 \rowcolor{lightcyan} \methodnamespace(ours) & ViT-B & 600& {K400} & \textbf{72.1} \\
    \hdashline
    OmniMAE ~\citep{girdhar2023omnimae} & ViT-B  & 800 & SSv2 &69.5 \\
    VideoMAE~\citep{tong2022videomae} & ViT-B & 800& {SSv2} &69.6  \\
    MGM~\citep{fan2023mgm} & ViT-B & 800& {SSv2} &70.6  \\
    MME~\citep{mme_sun} & ViT-B & 800& {SSv2} &70.0  \\
    SIGMA~\citep{salehi2025sigma} & ViT-B & 800 & {SSv2} &71.2  \\
    MGMAE~\citep{huang2023mgmae} & ViT-B & 800& {SSv2} &71.0  \\
 \rowcolor{lightcyan} \methodnamespace(ours) & ViT-B & 800& {SSv2} &\textbf{72.5} \\
 \bottomrule
    
\end{tabular}
\label{tab:ssv2}
\vspace{-10pt}
\end{table}
\vspace{-0.5cm}
\subsection{Ablations} \label{sec:ablation}
\vspace{-0.1cm}
We ablate various components of our \methodnamespace to show their impact on the downstream performance in Table~\ref{tab:ablation}. For computational efficiency, we use ViT-S as the backbone and pretrain with a smaller subset of K400, denoted as $\textrm{K400}_{\textrm{m}}$, containing around 80K samples. For evaluation, we  use $\textrm{K400}_{\textrm{m}}$ and $\textrm{SSv2}_{\textrm{m}}$ that denotes a subset of SSv2 with 50\% training data.  Unless stated otherwise, we use  CLIP features as the reconstruction target, add two synthetic objects,  mask 90\%, use trajectory masking, and train for 400 epochs.

\noindent\textbf{Synthetic motion, \textit{\underline{yes}}   or  \textit{no}?} 
In Table~\ref{tab:ablation_synthetic}, we show the impact of adding synthetic motions to the masked video modeling framework for both pixel and feature reconstruction. We can see that adding our synthetic motion patterns significantly improves the downstream performance of both pixel and feature reconstruction. 

\noindent\textbf{Reconstructing \textit{pixels} or \textit{\underline{features}}?} 
Table~\ref{tab:ablation_pixel_vs_features} compares pixel-based and feature-based reconstruction targets in masked video modeling. Reconstructing in feature space significantly boosts performance over pixel reconstruction, with a +12.1\% gain on spatially-focused tasks like $\textrm{K400}_{\textrm{m}}$ and +5.1\% on motion-focused tasks like $\textrm{SSv2}_{\textrm{m}}$. This suggests that using a high-level feature space as the reconstruction target provides more meaningful supervision for learning transferable spatio-temporal representations.

\noindent\textbf{Reconstructing which features: \textit{\underline{CLIP}},  \textit{MAE} or \textit{DINO}?} 
Table~\ref{tab:ablation_clip_vs_dino} compares several image foundation models—MAE~\citep{he2022masked}, DINO~\citep{caron2021emerging}, DINO-v2~\citep{oquab2023dinov2}, and CLIP~\citep{radford2021learning}—as reconstruction targets.  CLIP features outperform the others, likely due to CLIP’s image-text alignment training on a large dataset, enhancing its transferability and adaptability to video tasks as shown in ~\citep{wang2021actionclip,rasheed2023fine,ni2022expanding}. Compared to pixel reconstruction in Table~\ref{tab:ablation_pixel_vs_features}, all feature-based approaches show substantial performance gains.

\noindent\textbf{Trajectory-based masking: 
 \textit{\underline{yes}} or \textit{no?}}  
Table~\ref{tab:ablation_which_masking} shows the impact of our trajectory-based masking.  
Without trajectory masking, all tokens are masked using tube masking, which masks the whole video without distinguishing between object and non-object space-time tokens. In contrast, trajectory masking first masks space-time tokens associated with objects, followed by tube masking on the remaining tokens, maintaining the overall masking ratio. Our trajectory masking improves performance by approximately 1\% on both $\textrm{K400}_{\textrm{m}}$ and $\textrm{SSv2}_{\textrm{m}}$, validating that masking along object trajectories enhances motion-aware representation learning.

\begin{table}[t!]
 \vspace{-0.2cm}
\centering
\captionsetup[sub]{font=normalsize}
\caption{\textbf{Full finetuning comparison of various self-supervised methods on Kinetics-400}. All results are obtained with a ViT-B backbone. $*$ denotes results obtained by our evaluation. Our \methodnamespace achieves state-of-the-art performance. Top-1 accuracies (\%) are reported.}
\tablestyle{2.5pt}{1.02}
\small
\begin{tabular}{l|c|c|c|c}
    
    \toprule
    \textbf{Method} & \textbf{Backbone} & \textbf{Epochs} & \textbf{Pretrain} & \textbf{Top-1} \\
    SVT~\citep{svt} & ViT-B & - & K400 &78.4  \\
    VideoMAE~\citep{tong2022videomae} & ViT-B & 800& K400 &80.0 \\
    BEVT~\citep{wang2022bevt} & ViT-B & - & K400 &80.6 \\
    OmniMAE ~\citep{girdhar2023omnimae} & ViT-B  & 800 & K400 &80.8 \\
    MGM~\citep{fan2023mgm} & ViT-B & 800& K400 &80.8 \\
    MGMAE~\citep{huang2023mgmae} & ViT-B & 800& {K400} &81.2 \\
    SIGMA~\citep{salehi2025sigma} & ViT-B & 800 & K400 &81.5 \\
    MME$^*$~\citep{mme_sun} & ViT-B &800& K400 &{81.5} \\
 \rowcolor{lightcyan} \methodnamespace(ours) & ViT-B & 600& K400 &\textbf{83.1} \\
 \bottomrule
    
\end{tabular}
\label{tab:k400}
\vspace{-10pt}
\end{table}

\begin{table*}[htbp]
    \centering
    \tablestyle{4 pt}{1.12}
    \begin{subtable}{.25\linewidth}
        \centering
        \begin{tabular}{llcc}
        \toprule
        Synth. &  Target& $\textrm{K400}_{\textrm{m}}$ &  $\textrm{SSv2}_{\textrm{m}}$\\
        \midrule
         w/o  & Pixels & 66.0 & 55.1 \\
         w/  & Pixels & \textbf{68.1} & \textbf{56.8} \\
         \hline
         w/o &  Features & 78.1 & 60.2 \\
         w/ &  Features & \textbf{78.9} & \textbf{61.6} \\  
        \bottomrule
        \end{tabular}
        \caption{\textbf{Synthetic motion, \textit{yes} or \textit{no}?} Yes. It improves performance for both pixel and feature targets. }
        \label{tab:ablation_synthetic}
    \end{subtable}%
    \hspace{0.2cm} 
    \begin{subtable}{.22\linewidth}
        \centering
        \begin{tabular}{lcc}
        \toprule
        Target &  $\textrm{K400}_{\textrm{m}}$ & $\textrm{SSv2}_{\textrm{m}}$ \\
        \midrule
         MAE~\citep{he2022masked} & 77.6 & 59.5 \\
         DINO~\citep{caron2021emerging} & 77.6 & 59.7 \\
         DINOv2~\citep{oquab2023dinov2} & 77.8 & 59.7 \\
         CLIP~\citep{radford2021learning} & \textbf{78.1} & \textbf{60.2} \\
        \bottomrule   
        \end{tabular} 
        \caption{\textbf{Reconstructing which features?}  CLIP features provide the best target. No motion is added.
        }   
        \label{tab:ablation_clip_vs_dino}
    \end{subtable}
    \hspace{0.2cm} 
    \begin{subtable}{.2\linewidth}
        \centering
        \begin{tabular}{lcc}
        \toprule
        Transform & $\textrm{K400}_{\textrm{m}}$ & $\textrm{SSv2}_{\textrm{m}}$ \\
        \midrule
         None & 77.0 & 60.8 \\
         Scaling & 78.4 & 61.3 \\
         Rotation & 78.2 & 61.4 \\
         Both  &\textbf{78.9} & \textbf{61.6} \\
        \bottomrule
        \end{tabular}
        \caption{\textbf{Which object \textit{transformations}?} Using both rotation and scaling works the best.} \label{tab:ablation_which_object_transformations}
    \end{subtable} %
    \hspace{0.2cm} 
    \begin{subtable}{.22\linewidth}
        \centering
        \begin{tabular}{lcc}
        \toprule
        Training & $\textrm{K400}_{\textrm{m}}$ & $\textrm{SSv2}_{\textrm{m}}$ \\
        \midrule
         Single &   78.9 & 61.6 \\
         Mix &   79.5 & 62.2 \\
         Progressive &  \textbf{79.8} & \textbf{62.5} \\
        \bottomrule
        \end{tabular}
        \caption{\textbf{Which training strategy?} Progressive training outperforms both for the same number of iterations.
        }   
        \label{tab:ablation_which_training}
    \end{subtable}  
    
    \begin{subtable}{.25\linewidth}
        \centering
        \begin{tabular}{llcc}
        \toprule
        Ratio &  Target& $\textrm{K400}_{\textrm{m}}$ & $\textrm{SSv2}_{\textrm{m}}$ \\
        \midrule
         95\% &  features & 77.7 & 60.4 \\
         90\% &  features & 78.9 & 61.6 \\
         80\% &  features & \textbf{80.3} & \textbf{62.0} \\
        \bottomrule
        \end{tabular}
        \caption{\textbf{How much to mask?} Masking 80\% trajectory and total tokens.
        }   
        \label{tab:ablation_how_much_masking}
    \end{subtable}  
    \hspace{0.2cm} 
    \begin{subtable}{.22\linewidth}
        \centering
        \begin{tabular}{lcc}
        \toprule
        \# Objects & $\textrm{K400}_{\textrm{m}}$ & $\textrm{SSv2}_{\textrm{m}}$ \\
        \midrule
         1 & 78.6 & 61.0 \\
         2 & \textbf{78.9} & \textbf{61.6} \\
         3 & 78.6 & 61.3 \\
        \bottomrule
        \end{tabular}
        \caption{\textbf{How many \textit{objects}?}  Adding two object motions is the best.  } 
        \label{tab:ablation_how_many_objects}
    \end{subtable} %
    \hspace{0.2cm} 
    \begin{subtable}{.215\linewidth}
        \centering
        \begin{tabular}{lcc}
        \toprule
        Masking  &  $\textrm{K400}_{\textrm{m}}$ & $\textrm{SSv2}_{\textrm{m}}$ \\
        \midrule
        w/o Trajectory  & 77.8 & 60.7 \\
        w/  Trajectory   & \textbf{78.9} & \textbf{61.6} \\
        \bottomrule
        \end{tabular}
        \caption{\textbf{Trajectory  masking, \textit{yes} or \textit{no}?} Yes. Both targets are features. 
        }   
        \label{tab:ablation_which_masking}
    \end{subtable}%
    \hspace{0.20cm} 
    \begin{subtable}{.22\linewidth}
        \centering
        \begin{tabular}{lcc}
        \toprule
        Target & $\textrm{K400}_{\textrm{m}}$ & $\textrm{SSv2}_{\textrm{m}}$ \\
        \midrule
         Pixels & 66.0 & 55.1 \\
         Features & \textbf{78.1} & \textbf{60.2} \\
        \bottomrule
        \end{tabular}
        \caption{\textbf{Reconstructing \textit{pixels} or \textit{features}?}  Reconstructing features is much better. No motion is added. } 
        \label{tab:ablation_pixel_vs_features}
    \end{subtable} %
\caption{\textbf{Ablation experiments of our proposed \methodname.}
The default setting uses clip reconstruction targets with two synthetic objects and a 90\%  masking ratio with trajectory-based masking. We also show full-scale ablations with ViT-B and K400 pretraining in \textbf{supplementary}.}\label{tab:ablation}
\end{table*}

\noindent\textbf{How much to mask?}  
Table~\ref{tab:ablation_how_much_masking} shows the impact of masking ratio $m$ on downstream performance. We mask $m$ of space-time tubes within the objects and $m$ of space-time tubes in the whole video with $m$ being $95\%,90\%,$ or $80\%$.
We observe the best performance with \underline{$m$=80\%}.

\noindent\textbf{How many synthetic objects?}   In  Table~\ref{tab:ablation_how_many_objects}, we analyze the impact of adding more object motions to the input video on downstream performance. We evaluate adding  $1,2$ and $3$ object motions and compare the results. We observe that adding \underline{2 objects} is better than using a single object and the performance saturates when adding more objects. 

\noindent\textbf{Which object transformations?}  
In  Table~\ref{tab:ablation_which_object_transformations}, we show the impact of object transformation on downstream performance. Both rotation and scaling improve the performance individually and \underline{combining them} achieves the best results. 

\noindent\textbf{Which training strategy?}  
Table~\ref{tab:ablation_which_training} compares our proposed training strategies. We also compare learning with the single learning strategy, where only augmented videos are used for training.  We observe that both mixed and progressive learning are better than learning from augmented videos only. \underline{Progressive} is better than mixed, most likely due to a reduced domain shift in the second stage where only original videos are used.

\begin{figure}[h]
    \vspace{-0.4cm}
    \centering
    \captionsetup{font=small,skip=1mm}
    \includegraphics[width=0.95\linewidth]{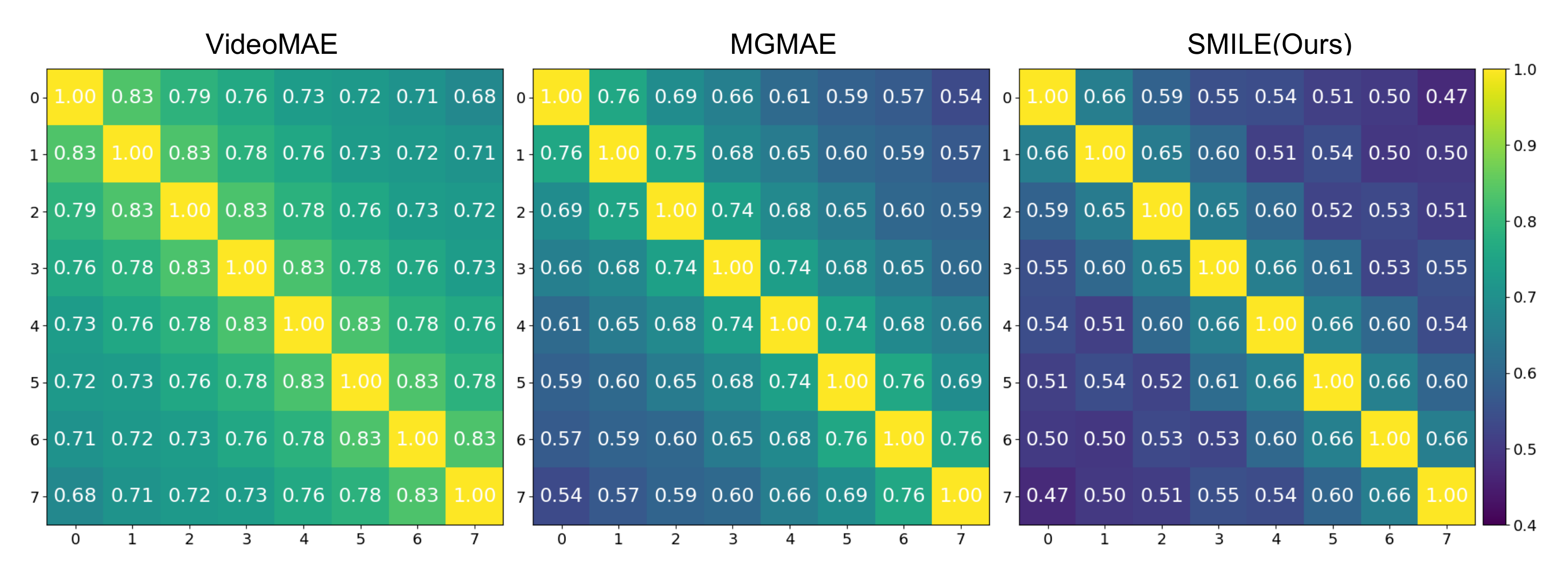}
    \caption{\textbf{Feature similarity across different frames for different SSL methods.}
    We compute this on K400 validation videos.} 
    \label{fig:sim_main}
\end{figure}

\noindent \textbf{Qualitative analysis.}
To compare the temporal dynamics of video representations learned by different SSL methods,  we compute their feature similarity across different frames of input videos (Figure~\ref{fig:sim_main}).  For our model, the features of different frames have larger differences, indicating better temporal awareness.  More plots are in the \textbf{supplementary}.

\begin{SCtable*}[][t!]
\setlength{\tabcolsep}{3pt}
\small
\begin{tabular}[t]{ll |cccc|cccc}
\toprule
\multicolumn{2}{c}{\textbf{Pretraining}} &\multicolumn{4}{c}{\textbf{Finetuning}} &\multicolumn{4}{c}{\textbf{Linear Probing}} \\
\midrule
\textbf{Method} & {Video Dataset} & {UCF} & {GYM} & {SSv2} & {EPIC} & {UCF} & {GYM} & {SSv2} & {EPIC} \\
\midrule
CLIP~\citep{radford2021learning} & - & 93.6 & 88.0 & 66.7 & 50.3 & 77.5 & 20.7 & 11.3 & 25.1 \\
ViCLIP~\citep{wang2023internvid} & Intervid-10M~\citep{wang2023internvid} & 95.2 & 89.7 & 67.9 & \underline{55.0} & \underline{86.7} & \underline{27.3} & \underline{18.9} & 27.3 \\
UMT~\citep{li2023unmasked} & K700~\citep{carreira2019short} & \underline{96.0} & \underline{89.9} & \underline{70.1} & 50.1 & \textbf{88.0} & 26.4 & 18.8 & \underline{28.2} \\
\midrule
\rowcolor{lightcyan} \methodnamespace(ours) & K400~\citep{Kinetics-400-arxiv} & \textbf{96.4} & \textbf{90.8} & \textbf{71.9} & \textbf{63.3} & 83.8 & \textbf{30.2} & \textbf{23.7} & \textbf{34.6} \\
\bottomrule
\end{tabular}
\caption{\textbf{Comparision with CLIP adaptations for video representation learning.} Despite using less data, we learn better video representations than other CLIP adaptations, especially for motion-focused downstream domains.}
\label{tab:clip_adaptations}
\end{SCtable*}

\begin{table*}[t!]
    \vspace{-0.1cm}
    \centering
        \caption{\textbf{Generalization assessment on SEVERE benchmark~\citep{thoker2022severe}} for recent video SSL methods in terms of domain shift, sample efficiency, action granularity, and task shift. We evaluate prior works with the official ViT-B backbone checkpoints pretrained on the K400 dataset. \methodnamespace shows the best generalization performance across setups. Adding synthetic motions further improves the performance.}
        
    \setlength{\tabcolsep}{5pt}
    \small
    \begin{tabular}{l ccccc cccc c}
    \toprule
    \multirow{2}{*}{\textbf{Method}} & \multicolumn{2}{c}{\textbf{Domain shift}} & \multicolumn{2}{c}{\textbf{Sample efficiency ($10^{3}$) }} & \multicolumn{2}{c}{\textbf{Action granularity}} & \multicolumn{2}{c}{\textbf{Task shift}} & \multirow{2}{*}{\textbf{Mean}} \\
    \cmidrule(lr){2-3} \cmidrule(lr){4-5} \cmidrule(lr){6-7} \cmidrule(lr){8-9} 
    & SSv2 &Gym99 & UCF & GYM & FX-S1 & UB-S1 & UCF-RC$\downarrow$ & Charades & \\
    \midrule
    VideoMAE~\citep{tong2022videomae}  & 68.6 & 86.6 & 74.6 & 25.9 & 42.8 & 65.3 & {0.172} & 14.4 & 57.6 \\
    MVD~\citep{mvd_wang}& {70.0} & 82.5 & 67.1 & 17.5 & 31.3 & 50.5 & 0.184 & 16.1 & 52.1 \\
    MGMAE~\citep{huang2023mgmae} & 68.9 & 87.2 & 77.2 & 24.1 & 33.7 & 79.5 & 0.181 & 17.9 & 58.8 \\
    MGM~\citep{fan2023mgm}&{71.1} & {89.1} & {78.4} & {26.4} & {38.6} & {86.9} & \textbf{0.152} & {22.5} & {62.2} \\
    SIGMA~\citep{salehi2025sigma} &{70.9} & {89.7} & {84.1} & {28.0} & \underline{55.1} & {79.9} & {0.169} & {23.1} & 64.2 \\
    MME~\citep{mme_sun}&{70.1} & {89.7} & {79.2} & {29.8} & \textbf{55.5} & {87.2} & \underline{0.155} & {23.6} & \underline{65.0} \\
    \rowcolor{lightcyan} \methodnamespace w/o motion &\underline{71.6} & \underline{90.0} & \underline{85.2} & \underline{32.0} & 53.4 & 74.3 & {0.175} & \underline{30.5} & {64.9} \\
    \rowcolor{lightcyan} \methodnamespace (ours)  &\textbf{72.1} & \textbf{90.8} & \textbf{86.4} & \textbf{35.1} & \underline{55.1} & \textbf{88.3} & {0.170} & \textbf{32.5} & \textbf{67.9} \\
    \bottomrule
    \end{tabular}

    \label{tab:severe-performance}
\vspace{-0.3cm}
\end{table*}

\vspace{-0.4cm}
\subsection{Generalization Assessment} \label{sec:severe}
\vspace{-0.1cm}
To demonstrate the generalization of our learned video representations to diverse downstream video setups, we evaluate \methodnamespace on the SEVERE benchmark~\citep{thoker2022severe}, which includes 8 experiments targeting 4 key generalization factors: \textit{domain shift}, \textit{sample efficiency}, \textit{action granularity}, and \textit{task shift}. 
Detailed configurations are provided in the \textbf{supplementary}. Beyond SEVERE, we also show generalization to more temporal tasks in \textbf{supplementary}.

Table~\ref{tab:severe-performance} shows the results. For the \textbf{Domain Shift}, we test cross-domain adaptability on SSv2 and Gym99, which differ from the pretraining dataset (K400). Our method significantly outperforms all others, demonstrating superior adaptability to domain shifts. For \textbf{Sample Efficiency}, we evaluate low-shot action recognition on UCF and GYM with only 1,000 samples for finetuning. Our approach surpasses previous methods, achieving a 5\% improvement on GYM over the best prior model, MME, showcasing its strong low-shot recognition capability even in motion-sensitive tasks.
For the \textbf{Action Granularity}, tested on FX-S1 and UB-S1 from FineGym, our model achieves the best results, excelling at distinguishing fine-grained motions within gymnastics routines (e.g., jump variations in FX-S1). It outperforms motion-aware models like MME, MGMAE, and MGM, highlighting superior motion generalizability. For \textbf{Task Shift}, evaluated with temporal repetition counting on UCF-RC~\citep{ucfrep-zhang2020context} and multi-label recognition on Charades~\citep{charades-sigurdsson:hal-01418216}, our method matches prior performance on repetition counting and achieves a 9\% improvement over MME on Charades, demonstrating broad task adaptability.
Overall, \methodnamespace (without synthetic motion) is on par with the prior best with the mean performance of 64.9\%. With synthetic motion \methodname, we observe a further 3.0\% gain, reinforcing our model’s generalization strength through synthetic motion.

\section{Discussions} \label{sec:discussion}
\vspace{-0.2cm}
\noindent\textbf{Comparison with Other CLIP Adaptations.}
Table~\ref{tab:clip_adaptations} presents a comparison between our \methodnamespace and the CLIP adaptation methods ViCLIP~\citep{wang2023internvid} and UMT~\citep{li2023unmasked}, along with the CLIP baseline.
Our method achieves superior performance on motion-sensitive datasets, particularly in linear probing, indicating that our CLIP adaptation learns better video representations than the video-text alignment of ViCLIP and the unmasked feature alignment of UMT, despite using less data.  More results are in the \textbf{supplementary}.

\begin{table}[t]
\centering
\tablestyle{2.0pt}{1.02}
\caption{\textbf{Learning without natural videos.} We train a ViT-S for pixel reconstruction.  Our synthetic object motions can learn video representations without using any natural videos. 
}   
\label{tab:types_of_synthetic_videos}
\small
\centering
\begin{tabular}{lccc}
\toprule
\textbf{Unnatural Video Scheme} &   Pretrain Data &$\textrm{K400}_{\textrm{m}}$ & $\textrm{SSv2}_{\textrm{m}}$\\
\midrule
 No Pretraining &   -& 39.7 & 28.2 \\
 \hdashline
 Video-frame  &  $\textrm{K400}_{\textrm{m}}$-1frm  &54.0 & 43.1 \\
 Video-frame + Motion&  $\textrm{K400}_{\textrm{m}}$-1frm  &63.6 & 53.2 \\
 \hdashline
 Image &  Places-80K  &52.1 & 40.5 \\
 Image + Motion &Places-80K &  60.1 & 52.1 \\
 \hdashline
 Black image + Motion &  -  & 59.2 & 50.0 \\
 Noise image + Motion &  - &  61.9 & 51.2 \\
 \bottomrule
\end{tabular}
\vspace{-0.4cm} 
\end{table}

\noindent\textbf{Learning without Natural Videos.}
In prior experiments, we overlay our object motions onto natural videos from the Kinetics-400 dataset, raising the question of whether downstream improvements stem from original video priors (motion, action, scene) or our added object motions. 
To investigate this, we construct videos by overlaying object motions onto four different backgrounds: (1) a single frame from natural videos, (2) a single scene image from the image dataset Places~\citep{zhou2017places}, (3) a plain black image, and (4) a noise image. Each background image is duplicated across $T$ frames, creating static clips onto which moving objects are placed. For (1), we select one frame per video from $\textrm{K400}_{\textrm{m}}$ (denoted $\textrm{K400}_{\textrm{m}}$-1frm), preserving action and scene cues without motion priors. For (2), we sample 80K images from Places (Places-80K), matching the dataset size of $\textrm{K400}_{\textrm{m}}$-1frm, to remove action and motion priors. Approaches (3) and (4) exclude all natural video or image priors entirely. Evaluation involves full finetuning on $\textrm{K400}_{\textrm{m}}$ and $\textrm{SSv2}_{\textrm{m}}$, consistent with our previous ablation studies.

Table~\ref{tab:types_of_synthetic_videos} reports results for different augmented video types.  Adding object motions consistently enhances downstream performance, whether combined with action and scene cues from natural videos (row 2 vs. row 3) or only scene cues from images (row 4 vs. row 5). Even with black or noisy backgrounds (rows 6 and 7), performance remains robust. Remarkably, using object motions alone yields substantial improvements—22.2\% on $\textrm{K400}_{\textrm{m}}$ and 23.0\% on $\textrm{SSv2}_{\textrm{m}}$ compared to no pretraining. These findings demonstrate our method’s effectiveness in learning video representations from synthetic motions alone. More results are in the \textbf{supplementary}. In summary, our method enables a new paradigm of learning strong video representations with masked video modeling without relying on natural videos.

\vspace{-0.2cm}
\section{Conclusions}
\vspace{-0.1cm}
In this work, we proposed \methodname, a novel self-supervised approach for video representation learning that addresses key limitations in existing masked video modeling methods by integrating both spatial and motion semantics. Our  \methodname, leverages the high-level spatial semantics from image-language pretrained models like CLIP and enhances motion representation through synthetic motion patterns. 
By capturing complex spatial features and motion dynamics, \methodname\ achieves superior performance on diverse downstream tasks, demonstrating improved discrimination and generalization over state-of-the-art video SSL methods. Our extensive evaluations demonstrated the advantages of our \methodname, setting a new paradigm for robust and effective video self-supervised learning. 

\paragraph{Acknowledgements.} This work is supported by the KAUST Center of Excellence for Generative AI under award number 5940. For computing time, this research used Ibex managed by the Supercomputing Core Laboratory at King Abdullah University of Science \& Technology (KAUST) in Thuwal, Saudi Arabia. We also thank Michael Dorkenwald for providing the synthetic objects used in this work.

{
    \small
    \bibliographystyle{ieeenat_fullname}
    \bibliography{main}
}
\clearpage
\appendix
\noindent\textbf{\Large{Appendix}}
\vspace{0.5 cm}

In this appendix, we present supplementary analysis and detailed experimental validations. Section~\ref{sec:detailed_k400_ssv2_comparision} provides an extensive state-of-the-art comparison for finetuning performance on K400 and SSv2 datasets. Section~\ref{sec:temporal_tasks} further evaluates the generalization capability of \methodnamespace on additional downstream video tasks. A detailed comparison against prior CLIP adaptation methods is provided in Section~\ref{sec:detailed_clip_comparision}. Section~\ref{sec:object_motion_only} shows more results for learning without natural videos. In Section~\ref{sec:add_exp}, we include additional experiments, such as performance under fixed training budgets, extended ablation studies with larger datasets, and qualitative visualizations. Finally, Section~\ref{sec:exp_details} specifies dataset details and clearly outlines the training and evaluation procedures used throughout our experiments.

\section{Extensive  Comparison on K400 and SSv2}
\label{sec:detailed_k400_ssv2_comparision}

In the main paper, due to the space limit,  we compare only with prior self-supervised methods using the same pretraining setup—specifically, a ViT-B backbone trained for 600 epochs on K400 and 800 on SSv2 datasets. Here, we provide a broader comparison, including self-supervised methods with varying pretraining setups and numerous supervised methods. Results for K400 and SSv2 finetuning are presented in Table~\ref{tab:k400_supplementary} and Table~\ref{tab:ssv2_supplementary}, respectively.

Our method not only reaches state of the art among all self-supervised methods but also matches or even surpasses many supervised methods 
that use specialized backbones, such as the hierarchical 3D transformer in MViTv2~\citep{li2022mvitv2} (more than +0.2\% on K400 and +1.6\% on SSv2) and the multi-head relation aggregation in Uniformer~\citep{li2022uniformer} (more than +0.1\% on K400, +0.9\% on SSv2), highlighting its ability to learn superior representations without heavily customized architectures. Additionally, we significantly outperform supervised methods using the same backbone (ViT-B), e.g. TimeSformer~\citep{Bertasius2021IsSA} by more than +2.4\% on K400 and +12.6\% on SSv2 and Mformer~\citep{patrick2021keeping} by more than +3.4\% on K400 and +5.4\% on SSv2. This demonstrates our method's strength in capturing better spatio-temporal dynamics than these supervised approaches.

Our approach significantly improves on VideoMAE~\citep{tong2022videomae} baseline, achieving gains of +3.6\% on K400 and +4.9\% on SSv2 for ViT-S, when trained for 800 epochs. Similarly, for ViT-B, it achieves an improvement of +3.1\% on K400 and +3.6\% on SSv2, when trained for 600 epochs on K400. Furthermore, our method outperforms all prior self-supervised approaches under similar pretraining settings, including the same backbone, dataset, and training epochs (e.g., K400 on ViT-B for 600/800 epochs).   This includes surpassing methods such as CV-MAE-V~\citep{cmae_v_lu}, which employs contrastive video masked autoencoding,  SIGMA~\citep{salehi2025sigma}, which uses Sinkhorn-Guided feature clustering, and OmniMAE~\citep{girdhar2023omnimae}, which reconstructs from both images and videos. These results underline the effectiveness of our method across diverse self-supervised learning strategies.  

The tables also compare our method with approaches that use significantly longer training schedules, such as MVD~\citep{mvd_wang}, ST-MAE~\citep{feichtenhofer2022masked}, and MotionMAE~\citep{yang2022motionmae}. MVD achieves strong performance but relies on a resource-intensive pipeline, involving 1200 epochs of pretraining for both VideoMAE~\citep{tong2022videomae} and MAE~\citep{he2022masked}, followed by 400 epochs of distillation, making direct comparisons challenging.   Despite training for only 600 epochs, our method surpasses MVD~\citep{mvd_wang} by 0.4\% on K400 finetuning and consistently outperforms other methods trained for 1600 epochs on K400 or 2400 epochs on SSv2. This includes motion-aware methods like MotionMAE~\citep{yang2022motionmae}, MME~\citep{mme_sun}, MGMAE~\citep{huang2023mgmae}, and MGM~\citep{fan2023mgm}, demonstrating the superior training efficiency of our approach. When trained for 1200 epochs \methodnamespace achieves a boost of  $0.3\%$ on both K400 and SSv2 finetuning showing the scalability of our method for longer training schedules.  

To summarize,  our method surpasses many supervised methods and achieves state-of-the-art performance among video SSL methods while maintaining training efficiency.

\begin{table}[t!]
\centering
\captionsetup[sub]{font=normalsize}
\caption{\textbf{Detailed comparison with supervised and self-supervised pretraining methods for full finetuning on  Kinetics-400 (K400)}. $*$ denotes results obtained by our evaluation. Params denote the number of parameters in millions. Our \methodnamespace outperforms many supervised methods, achieves the best performance among self-supervised methods, and demonstrates a faster convergence.}
\tablestyle{0.6pt}{1.02}
\small
\begin{tabular}{l|c|c|c|c|c}
    
    \toprule
    \textbf{Method} & \textbf{Backbone} & \textbf{Epochs} & \textbf{Pretrain} & \textbf{Top-1} & \textbf{{Params}} \\
    \shline
    \textit{supervised} & & & & & \\
    \hline
    Mformer~\citep{patrick2021keeping} &Mformer-B &{-} & K400 & 79.7 & 109 \\
    VideoSwin~\citep{liu2022video} & Swin-B & {-} & K400  &  80.6   & 88 \\
    TimeSformer~\citep{Bertasius2021IsSA} & ViT-B & {-} & K400 & 80.7 & 430 \\
    MViTv1~\citep{fan2021multiscale} & MViTv1-B & {-} & K400 & 80.2 & 37 \\
    MViTv2~\citep{li2022mvitv2} & MViTv2-B & {-} & K400 & 82.9 & 52 \\
    Uniformer-B~\citep{li2022uniformer} &Uformer-B &{-} & K400 & 83.0 & 50 \\
    \hline
    \textit{self-supervised} & & & & & \\
    \hline
    VideoMAE$^*$~\citep{tong2022videomae} & ViT-S & 800& K400 &75.9 &  22 \\
    VideoMAE~\citep{tong2022videomae} & ViT-S & 1600 & K400 &79.0 &  22 \\
  \rowcolor{lightcyan}  \methodnamespace(ours) & ViT-S & 800& K400 &\textbf{79.5} &  22 \\
     \hdashline
    VideoMAE~\citep{tong2022videomae} & ViT-B & 800& K400 &80.0 &  87 \\
    VideoMAE~\citep{tong2022videomae} & ViT-B & 1600& K400 &81.5 & 87 \\
    ST-MAE~\citep{feichtenhofer2022masked} & ViT-B & 1600 & K400 & 81.3 &  87 \\
    MVD~\citep{mvd_wang} &  ViT-B &  1600+400 &  K400 &  82.7 &   87 \\
    MotionMAE~\citep{yang2022motionmae} & ViT-B & 1600& K400 & 81.7 &  87 \\
    CMAE-V~\citep{cmae_v_lu} & ViT-B & 800 & K400 & 80.2 & 87 \\
    CMAE-V~\citep{cmae_v_lu} & ViT-B & 1600 & K400 & 80.9 & 87 \\
    BEVT~\citep{wang2022bevt} & ViT-B & 800+150 & K400 & 80.6 &  87 \\
    OmniMAE ~\citep{girdhar2023omnimae} & ViT-B  & 800 & K400 &80.8 & 87 \\
    SIGMA~\citep{salehi2025sigma} & ViT-B & 800 & K400 &81.5 &  87 \\
    MGM~\citep{fan2023mgm} & ViT-B & 800& K400 &80.8 &  87 \\
    MGM~\citep{fan2023mgm} & ViT-B & 1600& K400 &81.7 &  87 \\
    MME$^*$~\citep{mme_sun} & ViT-B &800& K400 &{81.5} &  87 \\
    MME~\citep{mme_sun} & ViT-B & 1600 & K400 & 81.8 & 87 \\
    MGMAE~\citep{huang2023mgmae} & ViT-B & 800& {K400} &81.2 &  87 \\
    MGMAE~\citep{huang2023mgmae} & ViT-B & 1600 & K400 & 81.8 &  87 \\
 \rowcolor{lightcyan} \methodnamespace(ours) & ViT-B & 600& K400 &\textbf{83.1} &  87 \\
 \rowcolor{lightcyan} \methodnamespace(ours) & ViT-B & 1200& K400 &\textbf{83.4} &  87 \\
 \bottomrule
    
\end{tabular}
\label{tab:k400_supplementary}
\vspace{-10pt}
\end{table}

\begin{table}[t!]
\centering
\captionsetup[sub]{font=normalsize}
\caption{\textbf{Detailed comparison with supervised and self-supervised pretraining methods for full finetuning on Something-Something V2 (SSv2)}. $*$ denotes results obtained by our evaluation. Params denote the number of parameters in millions. Our \methodnamespace outperforms many supervised methods, achieves the best performance among self-supervised methods, and demonstrates a faster convergence.  }
\tablestyle{0.6pt}{1.02}
\small
\begin{tabular}{l|c|c|c|c|c}
    \toprule
    \textbf{Method} & \textbf{Backbone} & \textbf{Epochs} & \textbf{Pretrain} & \textbf{Top-1} & \textbf{{Params}} \\
    \shline 
    \textit{supervised} & & & & & \\
    \hline
    Mformer~\citep{patrick2021keeping} & Mformer-B &{-} & K400 & 66.7 & 109 \\
    VideoSwin~\citep{liu2022video} & Swin-B & {-} & K400 & 69.6 & 88 \\
    TimeSformer~\citep{Bertasius2021IsSA} & ViT-B & {-} & K400 & 59.5 & 121 \\
    MViTv1~\citep{fan2021multiscale} & MViTv1-B & {-} & K400 & 67.7 & 37 \\
    MViTv2~\citep{li2022mvitv2} & MViTv2-B & {-} & K400 & 70.5 & 52 \\
    Uniformer-B~\citep{li2022uniformer} & Uformer-B &{-} & K400 & 71.2 & 50 \\
    \hline
    \textit{self-supervised} & & & & & \\
    \hline
    OmniMAE ~\citep{girdhar2023omnimae} & ViT-B & 800 & SSv2 & 69.5 & 87 \\
    VideoMAE~\citep{tong2022videomae} & ViT-B & 800 & {SSv2} & 69.6 & 87 \\
    VideoMAE~\citep{tong2022videomae} & ViT-B & 2400 & SSv2 & 70.8 & 87 \\
    CMAE-V~\citep{cmae_v_lu} & ViT-B & 800 & {SSv2} & 69.7 & 87 \\
    CMAE-V~\citep{cmae_v_lu} & ViT-B & 1600 & {SSv2} & 70.5 & 87 \\
    MME~\citep{mme_sun} & ViT-B & 800 & {SSv2} & 70.0 & 87 \\
    MGM~\citep{fan2023mgm} & ViT-B & 800 & {SSv2} & 70.6 & 87 \\
    MGM~\citep{fan2023mgm} & ViT-B & 2400 & {SSv2} & 72.1 & 87 \\
    SIGMA~\citep{salehi2025sigma} & ViT-B & 800 & {SSv2} & 71.2 & 87 \\
    MGMAE~\citep{huang2023mgmae} & ViT-B & 800 & {SSv2} & 71.0 & 87 \\
   {MGMAE~\citep{huang2023mgmae}} & ViT-B & 2400 & {SSv2} & 72.3 & 87 \\
 \rowcolor{lightcyan} \methodnamespace(ours) & ViT-B & 800& {SSv2} &\textbf{72.5}&  87 \\
    \hline
    VideoMAE$^*$~\citep{tong2022videomae} & ViT-S & 800 & K400 & 64.2 & 22 \\
    SIGMA~\citep{salehi2025sigma} & ViT-S & 800 & K400 & 68.7 &  22 \\
  \rowcolor{lightcyan}  \methodnamespace(ours) & ViT-S & 800& K400 &\textbf{69.1} &  22 \\
    \hdashline
    OmniMAE~\citep{girdhar2023omnimae} & ViT-B & 800 & K400 & 69.0 & 87 \\
    VideoMAE~\citep{tong2022videomae} & ViT-B & 800 & K400 & 68.5 & 87 \\
    MVD~\citep{mvd_wang} & ViT-B & 1600+400 & K400 & 72.5 & 87 \\
    MME~\citep{mme_sun} & ViT-B & 800 & {K400} & 70.5 & 87 \\
    SIGMA~\citep{salehi2025sigma} & ViT-B & 800 & {K400} & 71.1 & 87 \\
    MGMAE$^*$~\citep{huang2023mgmae} & ViT-B & 800 & {K400} & 68.9 & 87 \\
    MGM$^*$~\citep{fan2023mgm} & ViT-B & 800 & {K400} & 71.1 & 87 \\
 \rowcolor{lightcyan} \methodnamespace(ours) & ViT-B & 600& {K400} & \textbf{72.1}&  87 \\
 \rowcolor{lightcyan} \methodnamespace(ours) & ViT-B & 1200& {K400} & \textbf{72.4}&  87 \\
 \bottomrule
    
\end{tabular}
\label{tab:ssv2_supplementary}
\end{table}
\begin{table*}[t]
\centering
\setlength{\tabcolsep}{10pt}
\begin{tabular}{lcccccc}
\toprule
& \multicolumn{4}{c}{\textbf{Unsupervised Video Object Segmentation}} & \multicolumn{2}{c}{\textbf{Temporal Action Localization}}
\\
\cmidrule(lr){2-5} \cmidrule(lr){6-7}
 & \multicolumn{2}{c}{\textbf{Clustering}} & \multicolumn{2}{c}{\textbf{Overclustering}} &  \\
\cmidrule(lr){2-3} \cmidrule(lr){4-5} \cmidrule(lr){6-7}
\textbf{Method} & {YTVOS} & {DAVIS} & {YTVOS} & {DAVIS} & {THUMOS-14} & {ActivityNet-v1.3}\\
\midrule
VideoMAE~\citep{tong2022videomae} & 34.1 & 29.5 & 61.3 & 56.2 & 58.5 & 37.3 \\
MGM~\citep{fan2023mgm} & 36.6 & \textbf{36.5} & 61.2 & 56.6 & 62.0 & 37.6 \\
MGMAE~\citep{huang2023mgmae} & 34.5 & 31.0 & 60.1 & 57.5 & 56.3 & 37.3 \\
SIGMA~\citep{salehi2025sigma} & 37.5 & 31.5 & 66.4 & 58.5 &  62.7 &  37.7 \\
\rowcolor{lightcyan} \methodnamespace(ours) & \textbf{40.5} & {32.7} & \textbf{67.0} & \textbf{59.5}  & \textbf{65.6} & \textbf{38.0} \\
\bottomrule
\end{tabular}
\caption{\textbf{Generalization assessment on Unsupervised Video Object Segmentation and Temporal Action Localization.} 
All methods are evaluated on the ViT-B backbone pretrained on K400  with their publicly available checkpoints. \methodnamespace outperforms prior masked video modeling works on both tasks demonstrating a better temporal modeling capability for more complex video understanding tasks.}
\label{table:tal_un-vos}
\end{table*}

\section{Generalization to More Temporal Tasks} \label{sec:temporal_tasks}
In the main paper, we show the generalization capability of our method for diverse downstream settings in SEVERE-benchmark.  Here we show the generalization of our method to more video understanding tasks. In particular, we evaluate temporally aware tasks:  Unsupervised Video Object Segmentation (\textbf{Un-VOS}) and Temporal Action Localization (\textbf{TAL}). The goal is to evaluate the motion modeling capability of video representations with  \textbf{TAL} requiring motion boundary awareness and \textbf{Un-VOS} requiring object motion propagation modeling.

\subsection{Unsupervised Video Object Segmentation }
\noindent\textbf{Setup.} We adopt the evaluation approach from~\citep{salehi2025sigma} to assess the learned temporal and spatial features via unsupervised video object segmentation using the benchmark introduced by~\citep{salehi2023time}. Unlike conventional action recognition benchmarks that aggregate features into a single global representation, this task examines the encoder’s capability to generate consistent temporal object segmentation maps. Specifically, extracted space-time features are grouped using k-means clustering with a given number of clusters ($K$), then aligned to ground-truth object masks via the Hungarian algorithm~\citep{kuhn1955hungarian}. Segmentation accuracy is quantified through mean Intersection over Union (mIoU). The scenario is labeled as clustering when $K$ equals the actual object count and as overclustering when $K$ surpasses this number. We follow the implementation from ~\citep{salehi2025sigma} and report mIoU on 	\textbf{DAVIS}~\citep{pont20172017} and 	\textbf{YTVOS}~\citep{xu2018youtube}. 
 
\noindent\textbf{Results.}
 As shown in Table~\ref{table:tal_un-vos}, \methodnamespace  obtains the best segmentation performance across all settings except for DAVIS clustering where it is the second best. In particular, we outperform prior motion modeling methods MGM and MGMAE by 4\% and 6\% on YTVOS clustering and by 6\% and 7\% on YTVOS overclustering. Interestingly, we also beat SIGMA which explicitly clusters the reconstructed features via Sinkhornkoop clustering. This demonstrates the superior motion modeling capability of our approach over standard pixel reconstruction, motion-guided pixel reconstruction, and feature clustering reconstruction approaches.

\subsection{Temporal Action Localization}

\noindent\textbf{Setup.} 
Temporal action localization (TAL)~\citep{liu2025opentad,zhao2023re2tal,zhang2022actionformer} is a task that aims to identify categories of actions that occur in a video and to locate the start and end timestamps of all action instances. It requires the model to understand not only the spatial semantics within the frames but also the temporal dynamics across frame sequences to capture the action process. We evaluated our \methodnamespace as well as the comparative methods on two representative TAL benchmarks THUMOS-14~\citep{jiang2014thumos} and ActivityNet-v1.3~\citep{caba2015activitynet}. We used the pretrained models from each method as the backbones for video spatio-temporal feature extraction and finetuned them with the TAL method ActionFormer~\citep{zhang2022actionformer} on both datasets using the OpenTAD framework~\citep{liu2025opentad}. Following the common  practice in the TAL community, we report the average mean average precision (mAP) over various temporal intersection of union (tIoU) values,  i.e., 10 tIoU values [0.5, 0.55, 0.6, 0.65, 0.7, 0.75, 0.8, 0.85, 0.9, 0.95] for ActivityNet-v1.3 and 5 tIoU values [0.3, 0.4, 0.5, 0.6, 0.7] for THUMOS-14.

\noindent\textbf{Results.} 
As shown in Table~\ref{table:tal_un-vos}, our \methodnamespace achieves the highest average mAPs on both benchmarks. Specifically, on THUMOS-14, \methodnamespace outperforms some methods by a large margin \eg VideoMAE by 7\% and MGMAE by 9\%.  On the more challenging and largescale dataset ActivityNet-v1.3, it shows 0.3\% improvement over the second-best SIGMA. More notably, on ActivityNet-v1.3, the performance of \methodnamespace is on par with the state-of-the-art TAL performance, which relies on  fully supervised finetuning with labeled Kinetics-400 videos (not shown in the table). Overall, the results show that \methodnamespace generalizes better to more complex downstream video tasks than the current masked video modeling approaches.

\section{More comparisons with CLIP adaptations}
\label{sec:detailed_clip_comparision}

In this section, we provide further comparisons between our proposed \methodnamespace and existing CLIP adaptation methods for action recognition. Specifically, we consider two distinct categories of CLIP-based adaptations: \textit{\textbf{without intermediate pretraining}} and \textit{\textbf{with intermediate pretraining}}. The former approaches either jointly finetune the CLIP vision and text encoders using labeled video-text pairs of the target dataset (e.g., X-CLIP~\citep{ni2022expanding}, ViFi-CLIP~\citep{rasheed2023fine}, ILA~\citep{tu2023implicittemporalmodelinglearnable}) or directly finetune the CLIP vision encoder with added specialized spatio-temporal adaptation modules on labeled videos from the target dataset (e.g., AIM~\citep{yang2023aimadaptingimagemodels}, DUAL Path~\citep{park2023dualpathadaptationimagevideo}). In contrast, intermediate pretraining methods, such as UMT~\citep{li2023unmasked} and ViCLIP~\citep{wang2023internvid} use the CLIP model and intermediate video-text pairs for further pretraining to align video and text modality using contrastive learning.  Detailed results of these comparisons are presented in Table~\ref{tab:clip_adaptations_supp}.

We observe that \methodnamespace significantly outperforms prior adaptation methods that directly finetune the CLIP model on the target dataset. It achieves the best performance on all target datasets except K400, where it achieves comparable results. This highlights the effectiveness of our approach as a superior CLIP adaptation strategy which relies only on the unlabeled videos for adaptation. Moreover, as demonstrated in the main paper, \methodnamespace surpasses adaptation methods that also employ intermediate pretraining, UMT, and ViCLIP. Notably, the performance gains are particularly pronounced on motion-intensive datasets like GYM99, SSv2, and EPIC, underscoring the critical importance of explicit motion modeling an aspect often neglected in existing CLIP adaptations.  Our intuition is that CLIP features contain object information like shape, boundaries, and location, which guides the reconstruction task to focus on the overlaid objects as well as the original video semantics.

In summary, \methodnamespace provides a robust CLIP adaptation by reconstructing masked video inputs directly within the CLIP visual feature space while explicitly integrating synthetic motion cues.

\begin{table*}[t]
\centering
\setlength{\tabcolsep}{3pt}
\small
\resizebox{\textwidth}{!}{
\begin{tabular}[t]{ll |ccccc|cccc}
\toprule
\multicolumn{2}{c}{\textbf{}} 
&\multicolumn{4}{c}{\textbf{Finetuning}} &\multicolumn{4}{c}{\textbf{Linear Probing}} \\
\midrule
\textbf{Method} & {Intermediate Dataset} & {K400} & {UCF} & {GYM} & {SSv2} & {EPIC} & {UCF} & {GYM} & {SSv2} & {EPIC} \\
\midrule
CLIP~\citep{radford2021learning} & - & {81.8} & 93.6 & 88.0 & 66.7 & 50.3 & 77.5 & 20.7 & 11.3 & 25.1 \\
\hdashline
\textit{\textbf{Without intermediate pretraining}} & & & & & & & & \\
X-CLIP ~\citep{ni2022expanding}& - & {83.8} & {92.0} & {75.2} & {57.4} & {52.7} & {-} & {-} & {-} & {-} \\
ViFi-CLIP ~\citep{rasheed2023fine}& - & {83.9} & {94.6} & {81.5} & {48.6} & {48.9} & {-} & {-} & {-} & {-} \\
 AIM ~\citep{yang2023aimadaptingimagemodels}& - & {83.9} & {94.0} & \underline{90.3} & {66.4} & {58.4} & {-} & {-} & {-} & {-} \\
ILA ~\citep{tu2023implicittemporalmodelinglearnable}& - & \underline{84.0} & {94.2} & {82.7} & {65.0} & \underline{60.8} & {-} & {-} & {-} & {-} \\
DUAL-Path~\citep{park2023dualpathadaptationimagevideo} & - & \textbf{85.4} & {-} & {-} & {69.6} & {-} & {-} & {-} & {-} & {-} \\
\hdashline
\textit{\textbf{With intermediate pretraining}} & & & & & & & & \\
ViCLIP~\citep{wang2023internvid} & Intervid-10M~\citep{wang2023internvid} & {82.4} & 95.2 & 89.7 & 67.9 & {55.0} & \underline{86.7} & \underline{27.3} & \underline{18.9} & 27.3 \\
UMT~\citep{li2023unmasked} & K700~\citep{carreira2019short} & {81.7} & \underline{96.0} & {89.9} & \underline{70.1} & 50.1 & \textbf{88.0} & 26.4 & 18.8 & \underline{28.2} \\
\midrule
\methodnamespace(ours) & K400~\citep{Kinetics-400-arxiv} & {83.1} & \textbf{96.4} & \textbf{90.8} & \textbf{71.9} & \textbf{63.3} & 83.8 & \textbf{30.2} & \textbf{23.7} & \textbf{34.6} \\
\bottomrule
\end{tabular}
}
\caption{\textbf{More comparison with CLIP adaptations.} \methodnamespace learns better video representations than both types of CLIP adaptations: with and without intermediate pretraining. The performance gap is wider on motion-focused domains.}
\label{tab:clip_adaptations_supp}
\end{table*}

\section{Learning without Natural Videos}
\label{sec:object_motion_only}
In the main paper, we show that our method can learn video representations by overlaying object motions on clips from natural videos, single frames from natural videos, single natural images, or even black and noise images.   This raises the question about the effectiveness of using object motions alone and how they compare with learning from natural videos.
To answer this we compare the performance of learning from object motions only with learning from natural video data. Specifically, we generate video clips by adding our synthetic object motions to randomly generated noise images. As in the main paper, we take a noise image, duplicate it $T$ times to form a static video clip, and then overlay objects with motion on top of it. We compare its performance with learning from natural videos of the K400  dataset. We use  ViT-S and ViT-B backbones for this experiment and results are shown in Table~\ref{tab:learning_without_natural_videos}.

We observe that learning with such augmented videos significantly improves on no pretraining. Compared to the VideoMAE baseline which uses 240K natural videos of K400, learning from our object motions only with pixel reconstruction shows a small gap in performance. This highlights the effectiveness of video representations learned only from the proposed object motions via unnatural videos created on the fly.
The gap is further reduced when feature reconstruction is used instead of pixel reconstruction, demonstrating the impact of using CLIP feature projections over raw pixels, even for such unnatural data. 
Overall, our proposed synthetic object motions can act as a strong supervisory signal in a standalone to learn video representations with masked video modeling.  We leave the scaling of such learning without natural videos for larger models to future works.

\begin{table}[t!]
\tablestyle{2.0pt}{1.02}
\caption{\textbf{Learning video representations with only object motions.} We train VideoMAE baseline on Kinetics-400 videos and ours with augmented clips generated by overlaying noise images with object motions. All settings train a ViT-S for 400 and ViT-B for 600 epochs. Our method trained without any natural videos lags only by a small margin compared to the VideoMAE baseline trained with natural videos from the Kinetics-400 dataset.}   
\label{tab:learning_without_natural_videos}
\centering
\setlength{\tabcolsep}{0.25em}
\small
\begin{tabular}{cccccc}
\toprule
Method & Data & Target& K400 & SSv2 & GYM \\
\midrule
\rowcolor{lightgray}{ViT-S} & & & & &  \\
 No Pretrain. &   - &  - &65.7 & 52.2  & 55.1 \\
 Ours   & Noise + Motion & Pixel&  72.7 & 59.1 & 72.5 \\
 Ours   & Noise + Motion & Features &  73.7 & 61.0 & 74.6 \\
 VideoMAE&  K400  & Pixel& {75.9} & {62.7} & {75.1}\\
\rowcolor{lightgray}{ViT-B} & & & & &  \\
 No Pretrain. &   - &  - &69.1 & 49.8  & 50.0 \\
 Ours   & Noise + Motion & Pixel&  74.5 & 60.0 & 77.2 \\
 Ours   & Noise + Motion & Features &  77.5 & 64.1 & 83.0 \\
 VideoMAE&  K400  & Pixel& 79.0 & 67.0 & 86.6 \\
 \bottomrule
\end{tabular}
\vspace{-0.4cm} 
\end{table}

\section{Additional Experiments}
\label{sec:add_exp}
\subsection{ Performance with fixed training budget }
\begin{figure}[h]
    \centering
    \captionsetup{font=small,skip=1mm}
    \includegraphics[width=\linewidth]{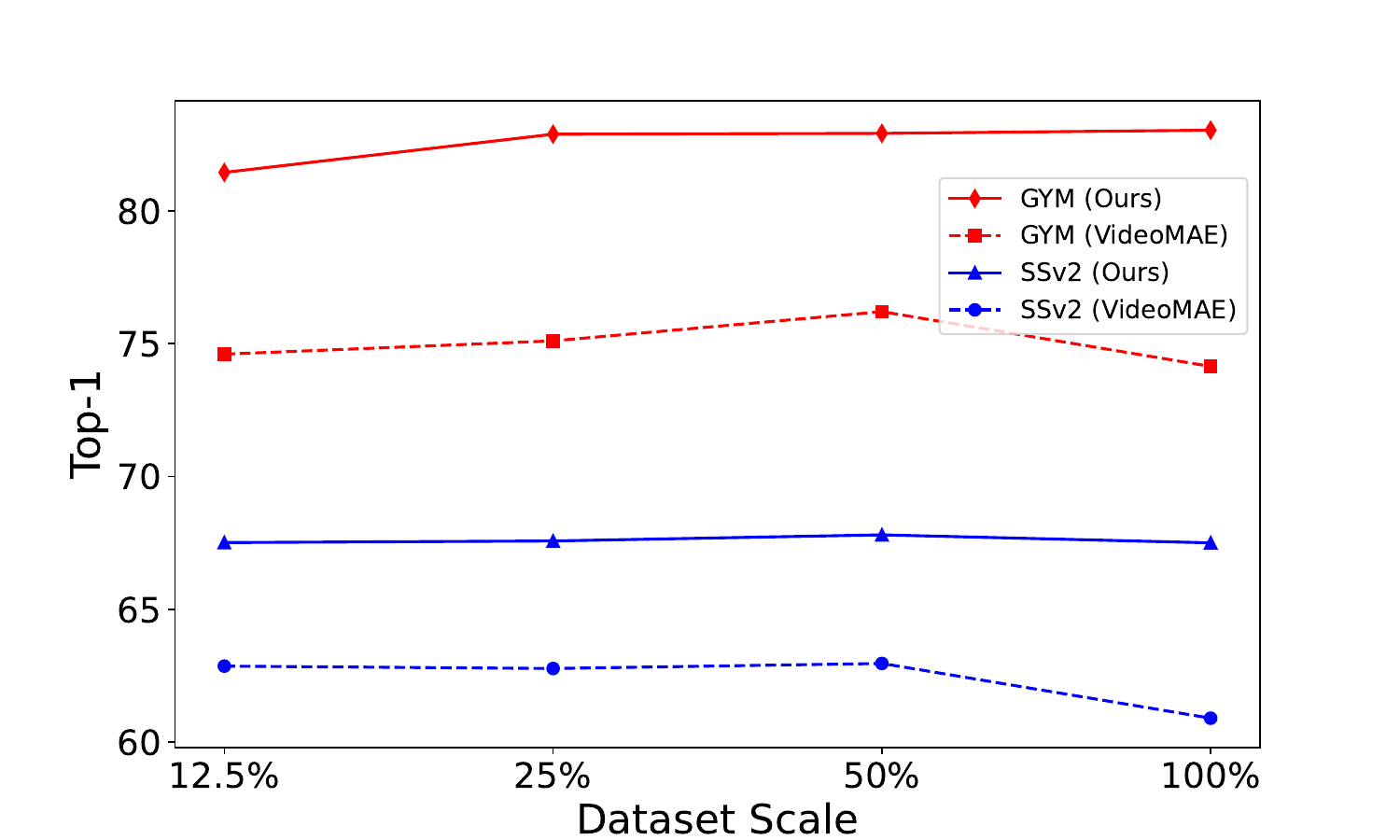}
    \caption{\textbf{Performance comparison with a fixed training budget.} We evaluate on SSv2 and GYM for full finetuning. Our method consistently outperforms VideoMAE~\citep{tong2022videomae} across all data scales with the same training budget.}
    \label{fig:budget}
\end{figure}

We now compare our method with VideoMAE~\citep{tong2022videomae} baseline
for different dataset scales with a fixed training budget, \textit{i.e.} total number of training iterations $n = s \times e$, where $s$ is the dataset size and $e$ is the number of epochs. Using the ViT-S backbone, we pretrain on four subsets of K400,  namely, 12.5\%, 25\%, 50\%, and 100\% of the full scale. For the whole dataset $s = 100\%$, we pretrain for 200 epochs; smaller subsets are trained for proportionally more epochs \textit{i.e.}, $s = 50\%$ for 400 epochs, $s = 25\%$ for 800, and $s = 12.5\%$ for 1600. Results in Figure~\ref{fig:budget} show that our method consistently outperforms the VideoMAE baseline by a large margin for various data scales under the same training budget. This highlights the robustness of our method to dataset size and training duration.

\begin{table}[h]
\tablestyle{2.0pt}{1.02}
\caption{\textbf{Full-Scale ablation.} Ablating our main contributions on a larger backbone and a bigger pretraining dataset, i.e., the original K400.  Reconstructing features and adding synthetic motions shows consistent improvements for a larger backbone (ViT-B) and scaling to a bigger pretraining dataset.}   
\setlength{\tabcolsep}{0.5em}
\centering
\begin{tabular}{ccccc}
\toprule
Backbone &  Target& Synth.& K400 & SSv2 \\
\midrule
 ViT-B &  Pixels &  w/o &78.3 & 67.2 \\
 ViT-B &  Pixels &  w/ &78.9 & 67.7 \\
 ViT-B &  Features &  w/o &81.2 & 70.8 \\
 ViT-B &  Features &  w/ &\textbf{81.7} & \textbf{71.2} \\
\bottomrule
\end{tabular}
\label{tab:ablation_full_scale}
\vspace{-0.4cm} 
\end{table}
\subsection{Full-scale ablation}
All main paper ablations are conducted with the smaller $\textrm{K400}_{\textrm{m}}$ pretraining and a ViT-S backbone. To reinforce the validity of our key contributions—feature target reconstruction and synthetic object motion, we extend the ablations to full-scale K400 pretraining using a larger ViT-B backbone. Full finetuning is performed on the complete K400 and SSv2 datasets. We adopt the best configurations from the small-scale ablations, including 80\% masking, trajectory masking, two object overlays, CLIP feature reconstruction, and a 300-epoch training schedule unless stated otherwise.

Table~\ref{tab:ablation_full_scale} presents the results for full-scale ablations. Consistent with the small-scale ablations in the main paper, incorporating synthetic motions boosts downstream performance. Specifically, our feature reconstruction improves the downstream performance by 2.9\% on K400 and 3.6\% on SSv2 over pixel reconstruction. By adding object motions, pixel reconstruction improves by 0.6\% on K400 and 0.5\% on SSv2, and feature reconstruction sees gains of 0.5\% on K400 and 0.4\% on SSv2. Our best configuration—feature reconstruction with synthetic motion—achieves the highest performance, reinforcing the robustness and scalability of our method across larger backbones and pretraining datasets.

\subsection{Qualitative analysis}
\begin{figure}[h]
    \centering
    \captionsetup{font=small,skip=1mm}
    \includegraphics[width=0.95\linewidth]{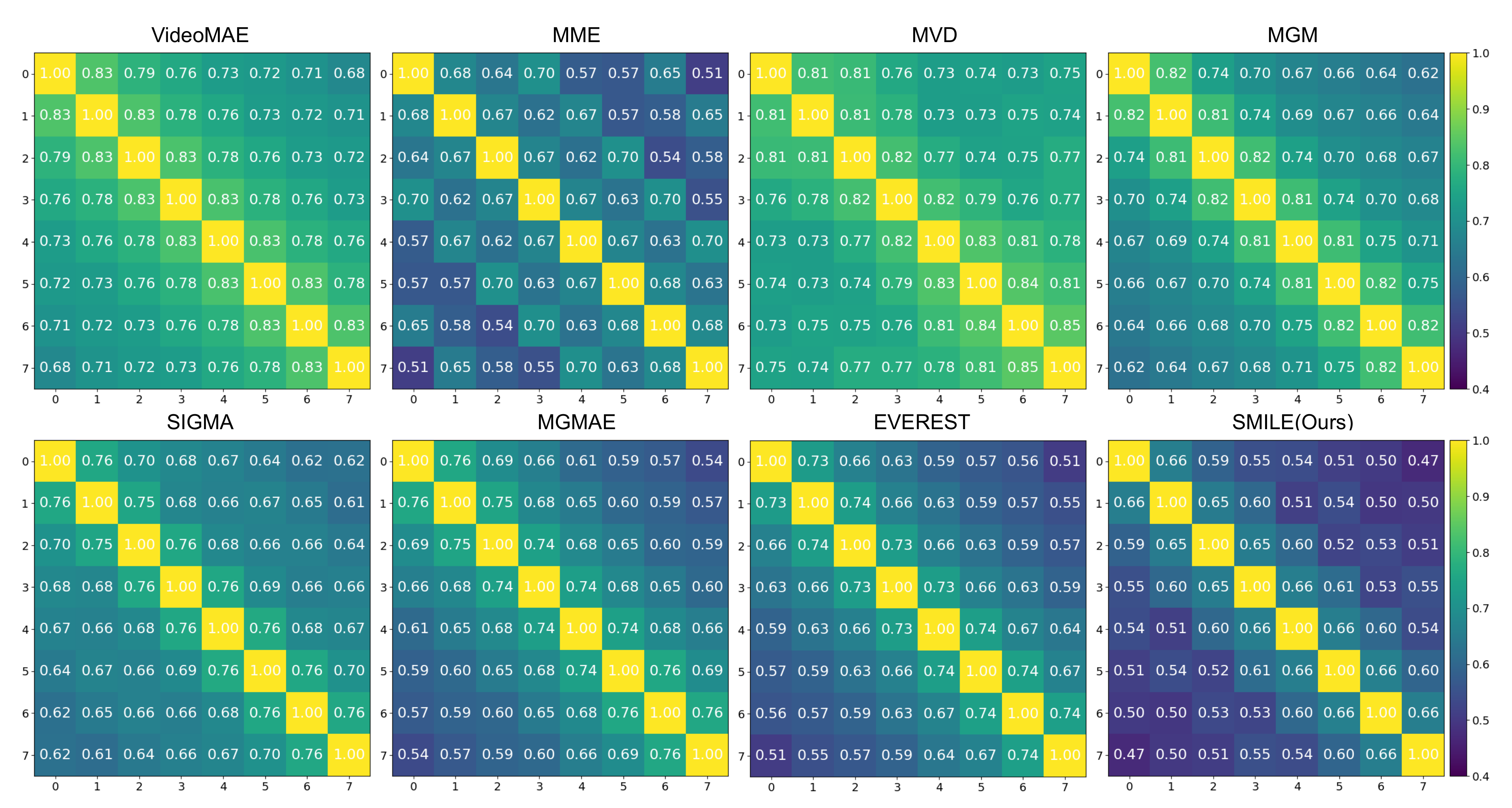}
    \caption{\textbf{Feature similarity across different frames for different SSL methods.}
    We compute this on K400 validation videos.} 
    \label{fig:sim_suppl}
\end{figure}

In Figure~\ref{fig:sim_suppl}, we extend the qualitative analysis to compare with more prior video SSL works. As before,  we observe that the features of different frames have larger differences for our model, indicating better temporal awareness. In particular, our feature similarity is consistent with moth motion-aware methods like MGM, MGMAE and MME demonstrating the motion focus of our method too.  

\begin{table*}[t]
    \centering
    \caption{\textbf{SEVERE benchmark.} Details of all the experimental subsets in the benchmark. We follow the configurations from the original work~\citep{thoker2022severe}.} 
    \label{tab:severe_desc}
    \setlength{\tabcolsep}{3pt}
    \small
    \begin{tabular}{llllrrrl}
    \toprule
    \textbf{Evaluation Setup} & \textbf{Experiment} & \textbf{Dataset} & \textbf{Task} & \textbf{\#Classes} & \textbf{\#Finetuning} & \textbf{\#Testing} & \textbf{Eval Metric}\\
    \midrule

     & Gym99 & FineGym~\citep{shao2020finegym} & Action Class. & 99 & 20,484 & 8,521 & Top-1 Acc.\\
    \midrule
    \multirow{2}{*}{\textbf{Sample Efficiency}}& UCF ($10^3$) & UCF 101~\citep{UCF-101-arxiv} & Action Class. & 101 & 1,000 & 3,783 & Top-1 Acc.\\
     & Gym ($10^3$) & FineGym~\citep{shao2020finegym} & Action Class. & 99 & 1,000 & 8,521 & Top-1 Acc.\\
    \midrule
    \multirow{2}{*}{\textbf{Action Granularity}} & FX-S1 & FineGym~\citep{shao2020finegym} & Action Class. & 11 & 1,882 & 777 & Mean-per-class\\
     & UB-S1 & FineGym~\citep{shao2020finegym} & Action Class. & 15 & 3,511 & 1,471 & Mean-per-class\\
    \midrule
    \multirow{2}{*}{\textbf{Task Shift}} & UCF-RC & UCFRep~\citep{ucfrep-zhang2020context} & Repetition Counting & - & 421 & 105 & Mean Error\\
     & Charades & Charades~\citep{charades-sigurdsson:hal-01418216} & Multi-label Class. & 157 & 7,985 & 1,863 & mAP\\
    \bottomrule
    \end{tabular}
\end{table*}

\section{Experimental Details}
\label{sec:exp_details}

\subsection{Datasets for main results}
For linear probing and full finetuning SoTA experiments,
we evaluate action recognition task with standard action recognition datasets \textit{i.e.} Kinetics-400~\citep{Kinetics-400-arxiv} (K400), SomethingSomething V2~\citep{goyal2017something} (SSv2), {UCF-101}~\citep{UCF-101-arxiv} (UCF), {HMDB-51}~\citep{kuehne2011hmdb} (HMDB), FineGYM~\citep{shao2020finegym} (GYM), and EPIC-Kitchens-100~\citep{EPIC-100-arxiv} (EPIC).  More details about the datasets are in Table~\ref{tab:dataset_details} and following:

\paragraph{Kinetics-400~\citep{Kinetics-400-arxiv}}
Kinetics-400 (K400) is a comprehensive benchmark designed for video action recognition tasks. It consists of over 306,000 concise video clips sourced from YouTube, spanning an impressive 400 distinct action categories. As one of the largest and most widely adopted datasets in this field, K400 plays a pivotal role in assessing and advancing cutting-edge models for understanding actions in video content.

\paragraph{SomethingSomething V2~\citep{goyal2017something}}
SomethingSomething V2 (SSv2) is a collaboratively sourced dataset consisting of first-person video recordings, specifically crafted to facilitate the development of common-sense reasoning capabilities. In terms of visual composition and perspective, it markedly diverges from Kinetics-400. The dataset comprises 168,913 training samples and 24,777 testing samples, distributed across 174 unique action categories.

\paragraph{UCF-101~\citep{UCF-101-arxiv}}
UCF-101 is a widely recognized benchmark in video self-supervised learning research. It comprises a diverse set of 9537 training and 3783 testing samples, sourced from YouTube videos grouped into 101 action categories, characterized by coarse granularity. Many of these categories show a substantial overlap with the action types included in Kinetics-400.

\paragraph{HMDB-51~\citep{kuehne2011hmdb}}
HMDB-51 (HMDB) is a widely-used benchmark for action recognition research. It features a total of 6,766 video clips, carefully selected from a variety of sources, such as films, the Prelinger Archive, YouTube, and Google Videos. The dataset is categorized into 51 unique action classes, with each class comprising no fewer than 100 video samples.

\paragraph{FineGYM~\citep{shao2020finegym}}
FineGYM (GYM) is a benchmark designed for fine-grained action analysis in gymnastics competitions. For our study, we specifically select the Gym-99 subset, which consists of 99 unique action categories. This subset provides 20,484 training samples and 8,521 testing samples.

\paragraph{EPIC-Kitchens-100~\citep{EPIC-100-arxiv}}
EPIC-Kitchens-100 (EPIC) is a large egocentric dataset capturing daily kitchen activities, annotated with 97 verbs and 300 nouns, where actions are defined as combinations of both. Similar to SS-v2, EK-100 differs significantly from Kinetics-400 in its unique visual style and first-person perspective. Using the standard splits provided in [13], it includes 67,217 training samples and 9,668 for validation, with our study focusing solely on recognizing the 97 verb classes.

\subsection{Datasets for SEVERE Benchmark}
\paragraph{SEVERE Benchmark\citep{thoker2022severe}}  
SEVERE-Benchmark spans eight experimental settings across four datasets \textit{i.e.} Something-Something V2, UCF, FineGYM, and Charades~\citep{charades-sigurdsson:hal-01418216}. Table~\ref{tab:severe_desc} provides detailed configurations for each subset.

\begin{table}[t]
    \centering
    \small
    \caption{\textbf{Datasets.} Details of the datasets used for evaluation showing the corresponding number of classes, training, and testing samples for each.}
     \begin{tabular}{lccc}
         \shline
         Dataset & \#Classes & \#Train & \#Test \\
         \shline
         K400 & 400 & 240K & 19K \\
         \midrule
         UCF & 101 & 9.5K & 3.8K \\
         HMDB & 51 & 4.8K & 2K \\
         SSv2 & 174 & 169K & 24.8K \\
         GYM & 99 & 20.5K & 8.5K \\
         EPIC & 97 & 67.2K & 9.7K \\
         \toprule
    \end{tabular}
    \label{tab:dataset_details}
\end{table}

\begin{table}[t]
    \centering
    \small
    \setlength{\tabcolsep}{3pt}
    \caption{\textbf{Linear-Evaluation setting.}}
    \label{tab:app-finetune-setting}
     \begin{tabular}{l|cccccc}
     \shline
    config & K400 & UCF & HMDB & SSv2 & GYM & EPIC \\
         \toprule
         optimizer & \multicolumn{6}{c}{AdamW\citep{adamw}} \\
         base learning rate & \multicolumn{6}{c}{1.e-3} \\
         weight decay & \multicolumn{6}{c}{0.05} \\
         optimizer momentum & \multicolumn{6}{c}{$\beta_1,\beta_2=0.9,0.999$} \\
         layer-wise lr decay~\citep{layer_wise} & \multicolumn{6}{c}{0.75} \\
         batch size & \multicolumn{6}{c}{128}\\
         learning rate schedule & \multicolumn{6}{c}{cosine decay} \\
         training epochs & 30 & 100 & 100& 50 &100 &100 \\
         flip augmentation & \emph{yes} & \emph{yes} & \emph{yes} & \emph{no} & \emph{yes} & \emph{yes} \\
         \bottomrule
    \end{tabular}
    \label{tab:linear-setting}
\end{table}

\begin{table}[t]
    \centering
    \small
    \caption{\textbf{Full finetuning evaluation setup.}}
    \label{tab:app-finetune-setting}
    \begin{tabular}{l|m{1cm}<{\centering}m{1cm}<{\centering}m{1cm}<{\centering}}
         \shline
         config & SSv2 & K400 & SEVERE \\
         \toprule
         optimizer & \multicolumn{3}{c}{AdamW} \\
         base learning rate & \multicolumn{3}{c}{1.0e-3} \\
         weight decay & \multicolumn{3}{c}{0.05} \\
         optimizer momentum & \multicolumn{3}{c}{$\beta_1,\beta_2=0.9,0.999$} \\
         layer-wise lr decay\citep{layer_wise} & \multicolumn{3}{c}{0.75} \\
         batch size & 32 & 16 & 16 \\
         learning rate schedule & \multicolumn{3}{c}{cosine decay} \\
         warmup epochs & \multicolumn{3}{c}{5} \\
         training epochs & 40 & 100 & 100 \\
         flip augmentation & \emph{no} & \emph{yes} & \emph{yes} \\
         RandAug~\citep{cubuk2019randaugment} & \multicolumn{3}{c}{(9,0.5)} \\
         label smoothing\citep{szegedy2015rethinking} & \multicolumn{3}{c}{0.1} \\
         mixup~\citep{zhang2018mixup} & \multicolumn{3}{c}{0.8} \\
         cutmix~\citep{yun2019cutmix} & \multicolumn{3}{c}{1.0} \\
         drop path & \multicolumn{3}{c}{0.1} \\
         \bottomrule
    \end{tabular}
    \label{tab:finetune-setting}
\end{table}

\begin{table}[h]
    \centering
    \small
    \caption{\textbf{Pretraining details.}}
    \label{tab:pretrain-setting}
    \begin{tabular}{l|m{1cm}<{\centering}m{1cm}<{\centering}m{1cm}}
         \shline
         config & SSv2 & K400\\
         \toprule
         optimizer & \multicolumn{2}{c}{AdamW} \\
         base learning rate & \multicolumn{2}{c}{1.5e-4} \\
         weight decay & \multicolumn{2}{c}{0.05} \\
         optimizer momentum & \multicolumn{2}{c}{$\beta_1,\beta_2=0.9,0.95$} \\
         batch size  & \multicolumn{2}{c}{256} \\
         learning rate schedule & \multicolumn{2}{c}{cosine decay} \\
         warmup epochs & \multicolumn{2}{c}{40} \\
         flip augmentation & \emph{no} & \emph{yes} \\
         augmentation & \multicolumn{2}{c}{MultiScaleCrop} \\
         Epochs & {800} & {600} \\
    \toprule
    \end{tabular}
    \label{tab:pretraining-setting}
\end{table}

\subsection{Training and Evaluation Details}
\noindent\textbf{Pretraining details.} We pretrain on Kinetics-400 (K400)~\citep{Kinetics-400-arxiv} and Something-Something V2 (SSv2)~\citep{goyal2017something}  datasets. To generate our segmented object set $O$,  we follow \citep{dorkenwald2024pin} to utilize Stable Diffusion~\citep{rombach2022high} and X-paste ~\citep{zhao2023x}, generating 60 samples for each of the 1203 categories in the LVIS dataset~\citep{gupta2019lvis}. Following  VideoMAE~\citep{tong2022videomae} we use a temporal stride of 2 for SSv2 and a stride of 4 for K400. Each clip contains 16 frames sampled at a resolution of \(224 \times 224\) pixels. Space-time tube embeddings are extracted using a 3D convolution layer, treating each \(2 \times 16 \times 16\) cube as a token. We use both tube and trajectory masking with a ratio $m=80\%$. 
We employ multiple sampling based on~\citep{hoffer2020augment} during the pretraining which effectively samples two input clips from the same video for reconstruction. This decreases the training time by almost half without any performance drop.
We always count epochs as “effective epochs $ = \textit{No. of epochs} \times \textit{No. of samples per video}$”, i.e., how many times each video is sampled and processed throughout training. We employ our progressive pretraining strategy for 600 epochs on K400 and 800 on SSv2,  300 and 400 in each stage, respectively. Table~\ref{tab:pretrain-setting} shows the rest of the configuration. We train our models with 8 NVIDIA V100 GPUs. For downstream evaluation, we only use the student network without its decoder and attach a task-dependent head to the pretrained student encoder e.g., a classification layer for action recognition.  

\noindent\textbf{Linear Probing details.} Table~\ref{tab:linear-setting} shows the settings for linear probing. We use 4 NVIDIA V100 GPUs for linear probing.

\noindent\textbf{Full Finetuning details.} Table~\ref{tab:app-finetune-setting} shows the settings for full finetuning,  following~\citep{tong2022videomae}. We use 4 NVIDIA V100 GPUS for fine-tuning.

\paragraph{SEVERE Benchmark evaluation.}  
We compare our method to recent masked video modeling approaches, using the SEVERE codebase~\citep{thoker2022severe} and keeping identical training and evaluation setups for fair comparison. Official models for each comparative method are used.

\end{document}